\documentclass[times, review, 10pt]{elsarticle}
\usepackage{algorithm}  
\usepackage{algorithmicx}  
\usepackage{algpseudocode} 
\usepackage{amsmath}
\usepackage{amssymb}
\usepackage{amsthm}
\usepackage{booktabs}
\usepackage{color}
\usepackage{caption}
\usepackage{dashrule}
\usepackage{epstopdf}
\usepackage{graphicx}
\usepackage{hyperref}
\usepackage[capitalise]{cleveref}
\usepackage{lineno}
\usepackage{multirow}
\usepackage{setspace}
\usepackage{subfigure}
\usepackage{tabularx}

\hypersetup{colorlinks = true}

\biboptions{sort&compress}
\begin{document}

\begin{frontmatter}
\title{Sparse and robust geometric twin support vector machine via asymmetric RoBoSS loss function}

\author[label1]{Kai Qi}\ead{qikai@cqnu.edu.cn}

\author[label2]{Xinji Huang}\ead{2024110510040@stu.cqnu.edu.cn}

\author[label2]{Hongchun Wang\corref{correspondingauthor}}\cortext[correspondingauthor]{Corresponding author. \emph{Email: wanghc@cqnu.edu.cn}}

\address[label1]{National Center for Applied Mathematics in Chongqing, Chongqing Normal University, Chongqing, 401331, China.}
\address[label2]{School of Mathematical Sciences, Chongqing Normal University, Chongqing, 401331, China.}

\begin{abstract}
In real-world scenarios, the training data usually contains redundant features, label noise and feature noise, which provide severe {challenges} for the efficiency of machine learning methods. Since { standard }support vector machine (SVM) adopts $l_2$-norm penalty and hinge loss function, it lacks the ability of selecting significant features and is sensitive to noise. To address these issues, this paper proposes a novel asymmetric, robust, bounded, sparse and smooth (aR) loss function for $l_1$-norm penalized geometric twin SVM (aRSGTSVM) to handle classification and regression tasks. The $l_1$-norm penalty can achieve the feature selection. The proposed aR loss function can not only effectively mitigate the impact of label noise, but also significantly enhance the stability to resampling noise, i.e., the zero-mean feature noise around the boundary hyperplanes. Furthermore, a statistical analysis of the robustness of aRSGTSVM was also conducted using the influence function. Since aRSGTSVM involves nonconvex and nonsmooth optimization, we develop a fast and stable proximal gradient descent based solving algorithm. Compared with related state-of-the-art methods, experimental results demonstrate the superiority of the proposed aRSGTSVM on both synthetic and UCI datasets. Furthermore, we apply aRSGTSVM to index tracking tasks, where results for tracking the different indices in the China stock market show that it can achieve satisfactory performance.
\end{abstract}

\end{frontmatter}

\section{Introduction}

Twin support vector machine (TSVM) \cite{Jayadeva2007twin} is one of the well-known variants of the support vector machine, which has experienced prosperous developments in the past decade and has been widely applied into different fields, such as digit recognition \cite{Chen2011Recursive}, medical diagnosis \cite{Qi2022Elastic,Liang2024Fuzzy} and bioanalysis \cite{Quadir2025Granular}. From a geometric perspective, TSVM aims to find a pair of nonparallel separating hyperplanes, where each hyperplane is required to be close to the samples of one class while being far from those of the other class. This makes TSVM more flexible and efficient than standard SVM. Moreover, similar to SVM, the standard TSVM can also be explained from a statistical point of view, i.e., it fits the ``loss + penalty'' regularization framework \cite{Shao2011Improvements,Qi2022Elastic}. Specifically, { most of the existing TSVM} adopt hinge loss and $l_2$-norm penalty. However, hinge loss is proved to be sensitive to {label noise} \cite{Wu2007Robust,Wang2024Fast} and {resampling noise} (zero-mean feature noise around the boundary hyperplanes) \cite{Huang2014Support}. The $l_2$-norm penalty lacks the ability to perform feature selection \cite{Zhu20041norm,Wang2006Doubly}. Therefore, there remains significant room for improving the efficiency of TSVM in handling complex real-world problems.

The label noise sensitivity of SVMs and TSVMs stems mainly from the upper-unbounded nature of the hinge loss. { Specifically, label noise usually lies near or even on the wrong side of the separating hyperplane. For such samples, the losses can be quite large due to the unboundedness of hinge loss function. Consequently, the final obtained separating hyperplane tends to be deviated by label noise by the minimization of the optimization objective. To address this issue, researchers have turned to nonconvex loss functions to enhance robustness against label noise.} By considering the difference between the logistic loss and its shifted version, Krause and Singer \cite{Krause2004Leveraging} proposed logistic difference (LD) loss for SVM to reduce the influence of label noise. Wu and Liu \cite{Wu2007Robust} integrated a truncated hinge loss, so-called ramp loss, with SVM to enhance the robustness. Later, Liu et al. \cite{Liu2015Ramp} combined the ramp loss with TSVM to propose robust nonparallel SVM (RNPSVM). Wang et al. \cite{Wang2019Robust} constructed a capped $l_1$-norm loss for robust TSVM. Wang et al. \cite{Wang2024Fast} designed a new truncated squared loss to obtain a robust SVM ($L_{tsl}$-SVM). In addition to truncation, correntropy is also an effective approach of designing robust losses \cite{Liu2007Correntropy,Singh2014Closs}. Xu et al. \cite{Xu2017Robust} developed the idea of correntropy to SVM and proposed rescaled hinge loss SVM, which can alleviate the disturbance of outliers. Xu et al. \cite{Xu2018Robust} studied the correntropy based loss (C-loss) for least squares SVM. Ma et al. \cite{Ma2021Adaptive} constructed a novel adaptive capped loss derived from correntropy to enhance TSVM's resistance to label noise. Recently, Akhtar et al. \cite{Akhtar2025RoBoSS} introduced a novel robust, bounded, sparse and smooth (RoBoSS) loss function for SVM, to mitigate the impact of label noise. {However, RoBoSS loss ignores the perturbation of resampling noise and the performance needs further improving.}

Regarding {resampling noise}, Huang et al. \cite{Huang2014Support} first demonstrated that hinge loss-based separating hyperplanes are sensitive to zero-mean feature noise near the boundary. Consequently, solutions derived from resampling procedures, such as $K$-fold cross-validation lack stability. Inspired by statistical quantiles, they introduced the pinball loss to SVM (PinSVM) to enhance its resampling stability. {The authors \cite{Huang2014Asymmetric} further extended this idea to squared loss, developing an asymmetric least squares SVM with a smooth, easily optimized objective function.} A key limitation of PinSVM is the loss of sample sparsity. Specifically, all training samples become support vectors. This drawback can increase the training burden \cite{Huang2014Support}. To encourage the sparsity, Shen et al. \cite{Shen2017Support} proposed a truncated pinball loss, which can provide a flexible trade-off between sparsity and feature noise insensitivity. Motivated by quantiles and correntropy, Yang and Dong \cite{Yang2019Robust} proposed a generalized quantile loss. Based on the least squares SVM, He et al. \cite{He2023Learning} established an asymmetric kernel-based learning classifier. 

For redundant features, they often provide irrelevant or even misleading information for the modelling process. As a result, the predicting performance may degrade. Zhu et al. \cite{Zhu20041norm} replaced $l_2$-norm penalty by $l_1$-norm penalty to propose 1-norm SVM, which can select significant features and remove redundant features, simultaneously. Ikeda and Murata \cite{Ikeda2005Geometrical} investigated the geometrical properties of $l_p$-norm penalized $\nu$-SVM, where $1\leq p\leq \infty$. Zhu et al. \cite{Wang2006Doubly} integrated elastic net with SVM to construct doubly regularized SVM (DrSVM). {In many real-world scenarios, input features exhibit a grouped structure. Consequently, there is greater interest in performing feature selection or elimination at the group level rather than on an individual basis.} To achieve group selection, Zou and Yuan \cite{Zou2008Fnorm} applied the infinity norm for each group of features to propose $F_{\infty}$-nrom SVM. Gao et al. \cite{Gao20111Norm} employed $l_1$-norm penalty for least squares TSVM to automatically select input features. Moosaei and Hlad\'{i}k \cite{Moosaei2023Sparse} proposed $l_p$-norm ($0<p<1$) least squares twin multi-class SVM. With $p\geq 1$, Xie et al. \cite{Xie2023Laplacian} designed Laplacian $l_p$-norm least squares TSVM.

{ Despite advances in robust SVM methods, existing approaches suffer critical limitations that hinder practical use. Label-noise-robust losses such as ramp loss, capped loss, and RoBoSS typically ignore resampling noise, while resampling-stable variants like pinball loss remain sensitive to label noise. Most robust losses either sacrifice sparsity or introduce nonconvexity with hyperparameters lacking theoretical tuning guidance. Moreover, few prior methods address label noise, resampling instability, and redundant feature interference simultaneously within a unified framework. Notably, RoBoSS lacks resampling-noise resistance and its feature-selection performance remains unexplored. In this paper, to address the aforementioned issues, we propose a novel asymmetric RoBoSS loss-based sparse geometric TSVM (aRSGTSVM). The proposed asymmetric RoBoSS loss is smooth and bounded, which can mitigate the impact of label noise and resampling noise, simultaneously. Moreover, aRSGTSVM incorporates an $l_1$-norm penalty to enable feature selection.} 

In short, the main contributions are summarized as follows:

1) We propose a new asymmetric and robust (aR) loss function. Especially, compared with RoBoSS loss function, the proposed aR loss function can not only effectively mitigate the impact of {label noise}, but also enhance the stability to the zero-mean noise near the boundary hyperplanes (so-called resampling noise).

2) We employ influence function to demonstrate the robustness of aR loss function. Although there is extensive literature on enhancing SVM robustness by constructing non-convex functions, few studies theoretically elucidate the robustness of the constructed functions. From a statistical perspective, we further proved that the influence function of aR loss is bounded, thereby providing a theoretical guarantee of its robustness.

3) Based on the aR loss and $l_1$-norm penalty, the aRSGTSVM models are proposed for robust classification and regression learning problems (In the following content, we refer to the regression model as aRSGTSVR to avoid confusion). In addition to robustness, the proposed aRSGTSVM and aRSGTSVR can also reduce the influence of noise features. 

4) To optimize the nonconvex and nonsmooth aRSGTSVM and aRSGTSVR optimization problems, we design an efficient algorithm based on the proximal gradient descent algorithm, termed as iPiano algorithm. The implemented algorithm is demonstrated to be fast and stable, and is applicable to high-dimensional learning problems. 

{5) A lot of numerical studies on synthetic and benchmark datasets demonstrate that the proposed aRSGTSVM and aRSGTSVR are robust against label noise and resampling noise, while maintaining excellent feature selection capability in high-dimensional settings.

6) To further validate the generalization capability, we apply the model into index tracking task. The results show that aRSGTSVR delivers consistently superior performance across different underlying indices.}

The rest of paper is organized as follows: Section \ref{sec:pre} introduces the related work of this paper. Section \ref{sec:main} proposes the aR loss function and applies it to both classification and regression problems. Section \ref{sec:num} presents a lot of experimental results of the proposed models on artificial datasets, UCI datasets and real-world stock datasets. Finally, in Section \ref{sec:con}, we revisit the main contributions of this paper and provide a concluding remark.

\section{Preliminaries and related work}\label{sec:pre}
First, we define some necessary notation. Consider a {supervised} learning data $\{(x_i^T, y_i)^T\}_{i=1}^n$ with $n$ samples and $p$ features, where $x_{i} = (x_{i1},x_{i2},\cdots,x_{ip})^T \in \mathbb{R}^p$ is the $i$-th sample. Note {that} $y_i \in \{ \pm 1\}$ means a binary classification problem, while $y_i\in \mathbb{R}$ means a regression problem. Furthermore, let $Y = \mathrm(y_{1},y_{2},\cdots y_{n})^T\in \mathbb{R}^n$ and $X = (x_{1},x_{2},\cdots x_{n})^T\in \mathbb{R}^{n\times p}$. {The positive and the negative samples are reorganized as $X_+$ and $X_-$, respectively.} Unless otherwise specified, all the vectors mentioned below are in column-form { and the norm defaults to the $l_2$-norm.}

\subsection{ENNHSVM}
{For the standard TSVM, its training process does not fully consider the core requirement of comparing distances to each separating hyperplane in the prediction stage, leading to an inconsistency between the training and prediction processes. As \cite{shao2014nonparallel} pointed out, this inconsistency can possibly reduce the prediction performance, especially in the datasets with heteroscedastic noise.} 

Recently, Qi and Yang \cite{Qi2022Elastic} proposed a novel consistent ENNHSVM, which performs better than the standard TSVM. Specifically, ENNHSVM pursues two nonparallel hyperplanes by optimizing the following problems:
\begin{equation}
	\begin{alignedat}{2}
		&\min && \frac{c_1}{2} \left( \|X_{+} w_{+} + e_{+} b_{+}\|^{2} + \|X_{-} w_{-} + e_{-} b_{-}\|^{2} \right)\\
        &&&+ \frac{1}{2} \left( \|w_{+}\|^{2} + b_{+}^{2} + \|w_{-}\|^{2} + b_{-}^{2} \right) \\
		&&& + \frac{c_2}{2} \left( \|\xi_{+}\|^{2} + \|\xi_{-}\|^{2} \right) + c_3 \left( e_{+}^{T} \xi_{+} + e_{-}^{T} \xi_{-} \right) \\
		&\text{s.t.} && \left\{
		\begin{aligned}
			& X_{+} w_{+} + e_{+} b_{+} + X_{+} w_{-} + e_{+} b_{-} \geq e_{+} - \xi_{+}, \xi_{+} \geq 0,\\
			& X_{-} w_{-} + e_{-} b_{-} + X_{-} w_{+} + e_{-} b_{+} \leq \xi_{-} - e_{-}, \xi_{-} \geq 0,
		\end{aligned} \right.
		\label{ENNHSVM}
	\end{alignedat}
\end{equation}
{ where $w_{\pm}$ and $b_{\pm}$ are the normal vectors and intercepts of the positive and negative hyperplanes. $c_i \ (i=1,2,3)$ are positive tuning parameters}, $\xi_{\pm}$ are slack variables and ${e}_{\pm}$ are two vectors with all elements equaling to one.

{ After obtaining $w_{\pm}$ and $b_{\pm}$, for a given new sample $x_{\rm new}$, its label can be classified via the following decision function:
\begin{equation}
f\bigl(x_{\rm new}\bigr) = \mathrm{sign}\Bigl(x_{\rm new}^T\bigl(w_+ + w_-\bigr) + \bigl(b_+ + b_-\bigr)\Bigr),
\label{decision function}
\end{equation}
where $\mathrm{sign}(\cdot)$ denotes the sign function.}

{Although} ENNHSVM is consistent and demonstrated to be efficient, it cannot perform feature selection and exhibits poor performance in the presence of redundant variables. Moreover, its application in regression has not yet been explored in depth.

\subsection{RoBoSS loss for SVM}
To reduce the impact of label noise, recently, Akhtar et al. \cite{Akhtar2025RoBoSS} proposed a robust, bounded, sparse and smooth loss (RoBoSS). They incorporated the RoBoSS loss into the SVM framework, resulting in a novel robust RoBoSS-SVM. 

{In detail}, RoBoSS loss is defined as 
\begin{equation}
	L_{RoBoSS}(u) = 
	\begin{cases} 
		\lambda \big(1 - (au + 1)\exp(-au)\big), & u > 0, \\
		0, & u \leq 0,
	\end{cases}
\end{equation}
where $a$ is the shape parameter and $\lambda$ is the boundary parameter. The primal problem of RoBoSS-SVM can be formulated as
\begin{equation}
	\min_{w,b} \frac{1}{2} \| w \|^2 + \frac{C}{n} \sum_{i=1}^n L_{\text{RoBoSS}} \left( 1 - y_i \left( w^T x_i + b \right) \right).
\end{equation}
{where $w$ and $b$ are the normal vector and intercept of the separating hyperplane. $C>0$ is the tuning parameter.}

\section{The proposed aRSGTSVM(R)}\label{sec:main}
As discussed earlier, although the RoBoSS loss is robust to noise, it exhibits instability to resampling. In this chapter, an asymmetric version of the RoBoSS loss, termed aR loss, is proposed to address more complex noise in high-dimensional learning. Specifically, \Cref{loss} introduces the aR loss and investigates its theoretical properties. \Cref{TSVM} presents the aRSGTSVM method for classification problems, while \Cref{TSVR} proposes the aRSGTSVR model for regression scenarios. Finally, \Cref{iPiano-algorithm} provides the corresponding optimization algorithm.

\subsection{asymmetric RoBoSS loss}\label{loss}
Although the RoBoSS loss demonstrates considerable robustness to label noise, Huang et al. \cite{Xu2017Novel} point out that in many practical problems, beyond outliers, high-dimensional complex data may also be affected by other types of noise, such as resampling noise or zero-mean noise near the bounding hyperplane. However, {one-sided loss functions like} the hinge loss and RoBoSS loss struggle to effectively mitigate the impact of such noise, leading to unstable model performance and leaving room for improvement. Therefore, to further enhance the capability of the RoBoSS loss in handling complex data, we propose a new loss function named aR, which is defined as follows:
\begin{equation}
	L_{aR}(u) = 
	\begin{cases} 
		\lambda \big( 1 - (au + 1) \exp(-au) \big), & u > 0, \\ 
		\tau \lambda \big( 1 - (au^2 + 1) \exp(-au^2) \big), & u \leq 0,
	\end{cases}
	\label{aR loss}
\end{equation}
where the shape parameter \(a > 0\), the margin parameter \(\lambda > 0\), and the parameter \(\tau \in [0,1]\) control the asymmetry of the loss function, thereby enhancing the model's robustness to resampling noise.

Next, we elaborate on the properties of the proposed aR loss function.

\textbf{Property 1.} The aR loss function is { $C^1\text{-smooth}$}.

\textbf{Proof.} {
First, by \eqref{aR loss}, it follows that
\begin{equation}
	\nabla L_{aR}(u) = 
	\begin{cases} 
		\lambda a^2 u \exp(-au), & u > 0, \\ 
		2\tau \lambda a^2 u^3 \exp(-au^2), & u \leq 0,
	\end{cases}
	\label{derivative-aR}
\end{equation}
Thus, we can deduce that \( \nabla L_{aR}(0^+) = 0 \) and \( \nabla L_{aR}(0^-) = 0 \). According to \eqref{aR loss}, we have \( \lim_{u \to 0^+} L_{aR}(u) = \lim_{u \to 0^-} L_{aR}(u) = L_{aR}(0) = 0 \), which implies that \( L_{aR}(u) \) is continuous at \( u = 0 \). Consequently, the aR loss is a { $C^1\text{-smooth}$} function.  $\square$
}

\textbf{Property 2.} The aR loss function is bounded.

\textbf{Proof.} { For \( u > 0 \), we have \( \nabla L_{aR}(u) = \lambda a^2 u \exp(-au) > 0 \), which implies that \( L_{aR}(u) \) is strictly monotonically increasing on \( (0, +\infty) \). For \( u < 0 \), we have \( \nabla L_{aR}(u) = 2 \tau \lambda a^2 u^3 \exp(-au^2) < 0 \), which implies that \( L_{aR}(u) \) is strictly monotonically decreasing on \( (-\infty, 0) \). According to the monotonicity above, the function reaches its global minimum at $u = 0$, and $ L_{aR}(0) = 0$, 
which means that the loss function is lower bounded by 0.

Next, we analyze the upper bound of the function by calculating the limits at infinity:
\begin{equation}
    \lim_{u \to +\infty} L_{aR}(u) = \lambda, \quad \lim_{u \to -\infty} L_{aR}(u) = \tau\lambda.
\end{equation}

Since \( 0<\tau\le1 \), we have \( \tau\lambda \le \lambda \). Combined with monotonicity on both intervals, all function values satisfy
\begin{equation}
    0 \le L_{aR}(u) \le \lambda, \quad \forall u\in\mathbb{R}.
\end{equation}

Therefore, the aR loss function is both lower bounded and upper bounded, namely, it is bounded on \( \mathbb{R} \).
$\square$}

{
\textbf{Property 3.} The aR loss is robust to label noise (outliers), which can be theoretically guaranteed by influence function.

\textbf{Proof.} Hampel introduced the influence function \cite{hampel1968contributions}, which is mainly used to measure the stability of an estimator when subjected to infinitesimal contamination. For an ideal roust loss function, the influence function of the induced estimator is bounded.

Following \cite{akhtar2026robots} and by \eqref{derivative-aR}, we have that $\nabla l_{aR}$ achieves its maximum at $u=\frac{1}{a}$ for $u>0$, while it achieves its maximum at $u=-\sqrt{\frac{3}{2a}}$ for $u\leq0$. Therefore, it follows that
\begin{equation}
    |\text{IF}(u)|\leq \max\left(\frac{\lambda a}{\text{e}}, \frac{3\tau \lambda \sqrt{6a}}{2\text{e}^{3/2}}\right)<\infty,\quad \forall u\in \mathbb{R}.
\end{equation}

Therefore, according to the influence function theory, for the SVM model designed based on the aR loss, even if the loss \(u\) caused by label noise is extremely large, their impact on the model is still limited. This theoretically guarantees the robustness of the aR loss.
}

\begin{figure}[!t]
	\centering
	\subfigure[Different $a$ values\label{subfig:a}]{
		\includegraphics[width=0.31\textwidth]{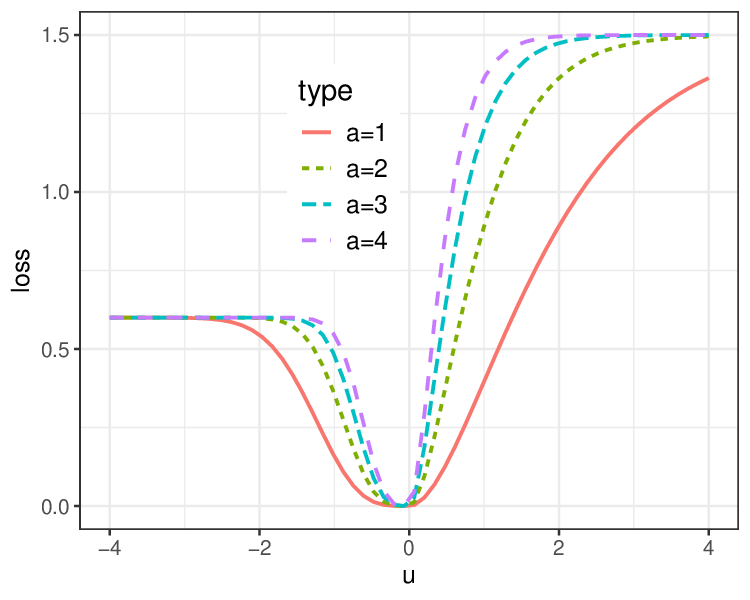}
	}\hfill
	\subfigure[Different $\lambda$ values\label{subfig:lambda}]{
		\includegraphics[width=0.31\textwidth]{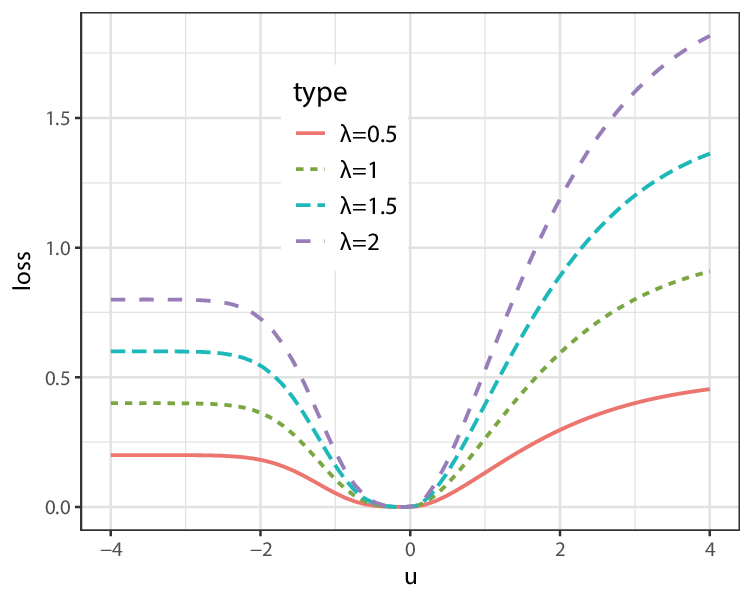}
	}\hfill
	\subfigure[Different $\tau$ values\label{subfig:tau}]{
		\includegraphics[width=0.31\textwidth]{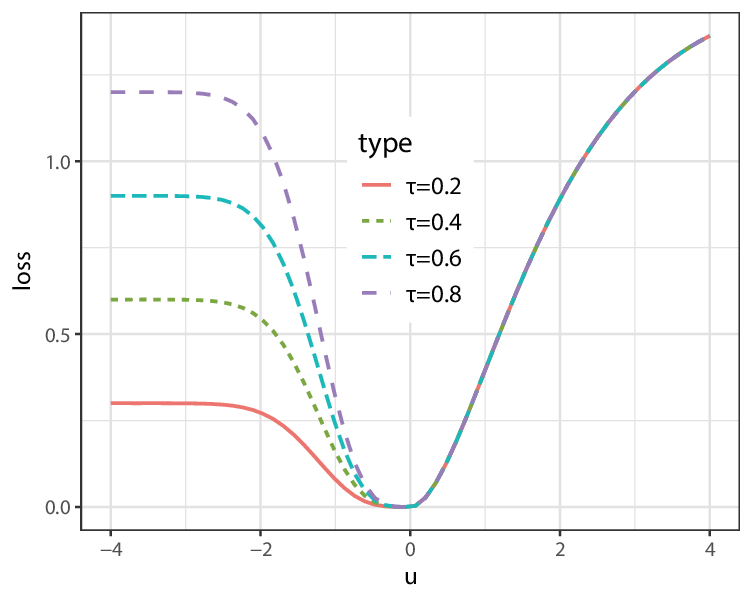}
	}
	\caption{Illustrations of different parameter values of the proposed aR loss function}
	\label{parameters}
\end{figure}

Figure \ref{parameters} illustrates the various forms of the aR loss function under different parameters. Combined with the preceding propositions, the asymmetric RoBoSS (aR) loss function is a bounded, asymmetric, smooth, and non-convex function. Consequently, compared to the original RoBoSS loss, our proposed loss function is not only robust to outliers but also resilient to zero-mean feature noise, enabling it to handle more complex problems.

\subsection{aR loss-based sparse geometric TSVM}\label{TSVM}
In this section, we mainly study the classification problems. Since the proposed aR loss can effectively mitigate the impact of outliers and strengthen the stability to resampling noise, simultaneously. We propose a novel sparse and robust geometric twin support vector machine(aRSGTSVM), which integrates the aR loss function with ENNHSVM, thereby inheriting the consistency properties of ENNHSVM. The resulting expression is as follows:

\begin{equation}
\begin{aligned}
\min \quad & \frac{c_1}{2}\left(\left\|X_{+} w_{+}+e_{+} b_{+}\right\|^{2}+\left\|X_{-} w_{-}+e_{-} b_{-}\right\|^{2}\right) \\
& +\lambda\left(\left\|w_{+}\right\|_{1}+\left|b_{+}\right| + \left\|w_{-}\right\|_{1}+\left|b_{-}\right|\right) \\
& +c_{2}{\left(\sum_{i\in N_+}L_{aR}\left((\xi_{+})_i\right)+ \sum_{i\in N_-}L_{aR}\left((\xi_{-})_i\right)\right)} \\
\text {s.t. } \quad &
\begin{cases}
X_{+} w_{+}+e_{+} b_{+}+X_{+} w_{-}+e_{+} b_{-} \geq e_{+}-\xi_{+}, \\
X_{-} w_{-}+e_{-} b_{-}+X_{-} w_{+}+e_{-} b_{+} \leq \xi_{-}-e_{-}, \\
\xi_{+} \geq 0, \xi_{-} \geq 0,
\end{cases}
\end{aligned}
\label{aRSGTSVM}
\end{equation}
{where $N_{\pm}$ denotes by the index sets of positive and negative samples. $(\xi_{\pm})_i$ means the $i$-th element of the slack variables. Other notations are similar to ENNHSVM.}

In contrast to ENNHSVM, our proposed classifier replaces the loss function in the original model with the aR loss and incorporates an $l_1$-norm penalty to equip the model with feature selection capability. The core idea of the proposed aRSGTSVM is outlined as follows:

\begin{itemize}
	\item[1)] Minimizing the first term of the objective of \eqref{aRSGTSVM} forces each hyperplane to be closer to its corresponding class samples and farther from the other class, enhancing inter-class discrimination.
	\item[2)] The second term \( \| cdot \|_1 \) induces sparsity in the model solution, enabling effective variable selection and handling of high-dimensional complex problems.
    \item[3)] The third term employs the aR loss function, whose inherent properties provide robustness against {label noise and resampling noise}, thereby improving the model's generalization performance.
\end{itemize}

To simplify subsequent derivations, define the following notations: $\tilde{X}_+ = (X_+, e_+)$, $\tilde{X}_- = (X_-, e_-)$, $\tilde{w}_{\pm} = (w_{\pm}^T, b_{\pm})^T$ and $\tilde{w} = (\tilde{w}_+, \tilde{w}_-)$. Furthermore, let 
\begin{equation}
A = 
\begin{pmatrix}
	\tilde{X}_+^T \tilde{X}_+ & 0 \\
	0 & \tilde{X}_-^T \tilde{X}_-
\end{pmatrix}, \quad B = 
\begin{pmatrix}
	\tilde{X}_+ & \tilde{X}_+ \\
	-\tilde{X}_- & -\tilde{X}_-
\end{pmatrix}, \quad e = 
\begin{pmatrix}
	e_+ \\
	e_-
\end{pmatrix}, \quad \xi = 
\begin{pmatrix}
	\xi_+ \\
	\xi_-
\end{pmatrix}.
\end{equation}

Then, the aRSGTSVM problem \eqref{aRSGTSVM} can be rewritten as:
\begin{equation}
	\begin{alignedat}{2}
		&\min && \frac{c_1}{2} \tilde{w}^T A\tilde{w} + \lambda \| \tilde{w} \|_1 + \frac{c_2}{2} \|\xi\|^2 \\
		&\text{s.t.} && \left\{
		\begin{aligned}
			& B\tilde{w} \geq e - \xi, \\
			& \xi \geq 0.
		\end{aligned} \right.
	\end{alignedat}
\end{equation}
which is equivalent to the following "loss + penalty" form, i.e.,
\begin{equation}
	\min_{\tilde{w}} \frac{c_1}{2} \tilde{w}^T A\tilde{w} + c_2 \sum_{i=1}^n L_{aR}(1 - b_i^T \tilde{w}) + \lambda \| \tilde{w} \|_1.
	\label{aRSGTSVM2}
\end{equation}
where $b_i$ represents the $i$-th row vector of matrix $B$.

\subsection{aR loss-based sparse geometric TSVR}\label{TSVR}
To extend the application scope from discrete classification to continuous prediction, we apply aRSGTSVM to regression problems and propose aR loss-based sparse geometric TSVR (aRSGTSVR).

Using the idea of TSVR \cite{khemchandani2016twsvr}, let $
X_{+} = \left( X_{+}, Y + \varepsilon_1 e_{+} \right),\ X_{-} = \left( X_{-}, Y - \varepsilon_2 e_{-} \right)$, $w_{+} = \left( w_{+}^T, \eta_{+} \right)^T, w_{-} = \left( w_{-}^T, \eta_{-} \right)^T $, {where $\varepsilon_1$, $\varepsilon_2 > 0$ are tuning parameters}. We first transform (\ref{aRSGTSVM}) into the following regression problem:
\begin{equation}
	\begin{aligned}
		\min\ & \frac{c_1}{2} \left( \|X_{+} w_{+} + \eta_{+} (Y + \varepsilon_{1} e_{+}) + e_{+} b_{+}\|^{2} + \|X_{-} w_{-} + \eta_{-} (Y - \varepsilon_{2} e_{-}) + e_{-} b_{-}\|^{2} \right) \\
		& + \lambda\left(\left\|w_{+}\right\|_{1}+\left|\eta_{+}\right|+\left|b_{+}\right| + \left\|w_{-}\right\|_{1}+\left|\eta_{-}\right|+\left|b_{-}\right|\right)\\
		& + c_{2}{\left(\sum_{i\in N_+}L_{aR}\left((\xi_{+})_i\right)+ \sum_{i\in N_-}L_{aR}\left((\xi_{-})_i\right)\right)}\\
		\text{s.t.}\ & 
		\begin{cases}
			X_{+} w_{+} + \eta_{+} (Y + \varepsilon_{1} e_{+}) + e_{+} b_{+} + X_{+} w_{-} + \eta_{-} (Y + \varepsilon_{1} e_{+}) + e_{+} b_{-} \geq e_{+} - \xi_{+}, \\
			X_{-} w_{-} + \eta_{-} (Y - \varepsilon_{2} e_{-}) + e_{-} b_{-} + X_{-} w_{+} + \eta_{+} (Y - \varepsilon_{2} e_{-}) + e_{-} b_{+} \leq \xi_{-} - e_{-} < 0, \\
			\xi_{+} \geq 0, \quad \xi_{-} \geq 0.
		\end{cases}
		\label{ENNHSVR}
	\end{aligned}
\end{equation}

Analogously to the classification problem, we further define the following variables:
$ \tilde{X}_+ = \left(X_+, Y + \varepsilon_1 e_+, 1\right),\ \tilde{X}_- = \left(X_-, Y - \varepsilon_2 e_-, 1\right),\ \tilde{w}_+ = \left(w_+^T, \eta_+, b_+\right)^T,\ \tilde{w}_- = \left(w_-^T, \eta_-, b_-\right)^T,\ $ $ \tilde{w} = \left(\tilde{w}_+^T, \tilde{w}_-^T\right)^T, $ as well as the corresponding matrices and vectors:
\[
A = \begin{pmatrix} \tilde{X}_+^T \tilde{X}_+ & O \\ O & \tilde{X}_-^T \tilde{X}_- \end{pmatrix},\ B = \begin{pmatrix} \tilde{X}_+ & \tilde{X}_+ \\ -\tilde{X}_- & -\tilde{X}_- \end{pmatrix},\ e = \begin{pmatrix} e_+ \\ e_- \end{pmatrix},\ \xi = \begin{pmatrix} \xi_+ \\ \xi_- \end{pmatrix}.
\] We rewrite (\ref{ENNHSVR}) as:
\begin{equation}
	\begin{alignedat}{2}
		&\min && \frac{c_1}{2} \tilde{w}^T A \tilde{w} + \lambda \| \tilde{w} \|_1 + \frac{c_2}{2} \| \xi \|^2 \\
		&\text{s.t.} && \left\{
		\begin{aligned}
			& B \tilde{w} \geq e - \xi, \\
			& \xi \geq 0.
		\end{aligned} \right.
		\label{ENNHSVR2}
	\end{alignedat}
\end{equation}

Similar to the previous steps, we obtain the linear case of the aRSGTSVR model:
\begin{equation}
	\begin{alignedat}{2}
		&\min && \frac{c_1}{2} \tilde{w}^T A \tilde{w} + c_2 \sum_{i=1}^n L_{aR} \left( 1 - b_i^T \tilde{w} \right) + \lambda \| \tilde{w} \|_1. 
		\label{aRSGTSVR}
	\end{alignedat}
\end{equation}

{ By utilizing the feature mapping $\boldsymbol{\Phi}(\cdot)$ that maps raw input space into a high dimensional Reproducing Kernel Hilbert Space, the proposed aRSGTSVM(R) model can be extended to the sparse kernel case. In fact, by only replacing $X$ with $\boldsymbol{\Phi}(X)$, we derive the aRSGTSVM(R) model for the sparse kernel scenario.} Note that feature selection is one of the key focuses of this study, so we primarily consider the linear case

\subsection{iPiano algorithm}\label{iPiano-algorithm}
Due to the non-convex aR loss function and non-smooth penalty term, Problem (\ref{aRSGTSVM}) constitutes a non-convex optimization problem. Conventional solution approaches typically employ the difference of convex functions (DC) algorithm or the half-quadratic (HQ) algorithm, which require iteratively solving a series of subproblems and incur substantial computational costs. Therefore, we opt to utilize the iPiano (Inertial Proximal Algorithm for Nonconvex Optimization) algorithm \cite{ochs2014ipiano} for optimization, which offers advantages of rapid convergence and numerical stability, particularly suited for high-dimensional complex problems. The following derivation will be presented using the classification problem as an illustrative example.

\subsubsection{iPiano for aRSGTSVM}
The iPiano algorithm can solve optimization problems of the form:
\begin{equation}
	\min_{x \in \mathbb{R}^n} h(x) = f(x) + g(x),
	\label{iPiano}
\end{equation}
where $g$ is convex (possibly nonsmooth), $f$ is { $C^1\text{-smooth}$} (possibly nonconvex).

Thus,  (\ref{aRSGTSVM}) can also be interpreted as having the form "$f + g$" :
\begin{equation}
	\begin{alignedat}{2}
		&\min_{\tilde{w}} && \underbrace{\frac{c_1}{2} \tilde{w}^T A \tilde{w} + c_2 \sum_{i=1}^n L_{aR} \left( 1 - b_i^T \tilde{w} \right)}_{f} + \underbrace{\lambda \| \tilde{w} \|_1}_{g}. 
		\label{aRSGTSVM3}
	\end{alignedat}
\end{equation}

Let \( u_i = 1 - b_i^T \tilde{w} \), then the objective function \( f \) is given by:
\begin{equation}
	f(\tilde{w}) = 
	\begin{cases} 
		\frac{c_1}{2} \tilde{w}^T A \tilde{w} + c_2 \sum\limits_{i=1}^n \lambda \left( 1 - \left( a u_i + 1 \right) \exp\left( -a u_i \right) \right), & u_i > 0, \\ 
		\frac{c_1}{2} \tilde{w}^T A \tilde{w} + c_2 \sum\limits_{i=1}^n \tau \lambda \left( 1 - \left( a u_i^2 + 1 \right) \exp\left( -a u_i^2 \right) \right), & u_i \leq 0. 
	\end{cases}
	\label{eq:piecewise_f}
\end{equation}

Then \( \nabla f \) is obtained:
\begin{equation}
	\nabla f(\tilde{w}) = 
	\begin{cases} 
		c_1 A\tilde{w} - c_2 \sum\limits_{i=1}^n a^2 \lambda b_i u_i \exp(-a u_i), & u_i > 0, \\ 
		c_1 A\tilde{w} - c_2 \sum\limits_{i=1}^n 2a^2 \tau \lambda b_i u_i^3 \exp(-a u_i^2), & u_i \leq 0. 
	\end{cases}
\end{equation}

{ Here, $L>0$ denotes an upper bound constant for the Lipschitz continuity of $\nabla f$, which satisfies the inequality
}

\begin{equation}
	|\nabla f(\tilde{w}_1) - \nabla f(\tilde{w}_2)| \leq L \|\tilde{w}_1 - \tilde{w}_2\|, \quad \forall \tilde{w}_1, \tilde{w}_2.
\end{equation}

Because
\begin{equation}
	\nabla^2 f(\tilde{w}) = 
	\begin{cases} 
		c_1 A - c_2 \sum_{i=1}^n a^2 \lambda b_i b_i^T \left( -1 + a u_i \right) \exp(-a u_i), & u_i > 0, \\ 
		c_1 A - c_2 \sum_{i=1}^n 2a^2 \tau \lambda b_i b_i^T \left( -3u_i^2 + 2a u_i^4 \right) \exp(-a u_i^2), & u_i \leq 0, 
	\end{cases}
\end{equation}
we can get
\begin{equation}\label{eq:L}
	L = c_1\| A \| +  c_2 \lambda a^2 \max\left(1, \frac{2 \left| \tau \right|}{a \sqrt{e}}\right) \sum_{i=1}^n \| b_i \|^2.
\end{equation}

The general framework of the iPiano algorithm for the problem (\ref{iPiano}) is presented in the algorithm \autoref{algorithm1}. { In this work, we empirically set the maximum iteration number $n_{\rm iter}$ to 500 and the error threshold $\varepsilon$ to $10^{-6}$.}

\begin{algorithm}[!t]
	\caption{iPiano for aRSGTSVM}
	\label{algorithm1}
	\begin{algorithmic}[1]
		\State \textbf{Input:} Set step size parameters $\beta=0.5$, $\alpha=\dfrac{1-\beta}{L}$, where $L$ is the Lipschitz constant of $\nabla f$, choose $\tilde{w}^0 \in \text{dom}, h$, $\tilde{w}^{-1} = \tilde{w}^0$. {The maximal iteration number $n_{iter}$, and the convergent error $\varepsilon$.}
		\State \textbf{Output:} Optimal solution $\tilde{w}^*$ of (\ref{aRSGTSVM})
		\State Initialize iteration counter $n = 0$
		\While {$n < n_{iter}$}
		\State $\tilde{w}^{n+1} = (I + \alpha_n \partial g)^{-1} \big(\tilde{w}^n - \alpha_n \nabla f(\tilde{w}^n) + \beta_n (\tilde{w}^n - \tilde{w}^{n-1})\big)$
		\If {$\|\tilde{w}^{n+1} - \tilde{w}^{n}\|_2 < \varepsilon$}
		\State \textbf{break}
		\EndIf
		\State $n = n + 1$
		\EndWhile
		\State \textbf{return} $\tilde{w}^{n+1}$
	\end{algorithmic}
\end{algorithm}

The proximal map is defined by
\begin{equation}
	(I + \alpha \partial g)^{-1}(\hat{x}) := \mathop{\mathrm{arg\,min}}\limits_{x \in \mathbb{R}^n} \left\{ \alpha g(x) + \frac{1}{2} \lVert x - \hat{x} \rVert_2^2 \right\},
\end{equation}
{ where $I$ is the identity map, $\alpha > 0$ is a given step size parameter, and $g$ is a proper lower semicontinuous convex function whose specific definition is given in (\ref{aRSGTSVM3}).}

\subsubsection{iPiano for aRSGTSVR}

According to \eqref{aRSGTSVR} and \eqref{aRSGTSVM2}, the aRSGTSVR model can be reformulated into a form consistent with that of the aRSGTSVM. Thus, similarly to algorithm \autoref{algorithm1}, the overall procedure for solving the aRSGTSVR algorithm is shown in algorithm \autoref{algorithm2}.

{ It is worth noting that the convergence of the proposed iPiano-based algorithm is guaranteed by Theorem 4.8 in the iPiano framework established by Ochs et al. \cite{ochs2014ipiano}.
Specifically, for nonconvex and possibly nonsmooth objective functions, under the condition that the objective satisfies the Kurdyka–Lojasiewicz property and appropriate step size parameters are adopted, it can be verified that the iterative sequence of the iPiano algorithm converges to critical points. Consequently, the algorithm developed has a solid theoretical convergence guaranty.}

\begin{algorithm}[!t]
	\caption{iPiano for aRSGTSVR}
	\label{algorithm2}
	\begin{algorithmic}[1]
		\State \textbf{Input:} Set step size parameters $\beta=0.5$, $\alpha=\dfrac{1-\beta}{L}$, where $L$ is the Lipschitz constant of $\nabla f$, choose $\tilde{w}^0 \in \text{dom}, h$, $\tilde{w}^{-1} = \tilde{w}^0$. {The maximal iteration number $n_{iter}$, and the convergent error $\varepsilon$.}
		\State \textbf{Output:} Optimal solution $\tilde{w}^*$ of (\ref{aRSGTSVR})
		\State Initialize iteration counter $n = 0$
		\While {$n < n_{iter}$}
		\State $\tilde{w}^{n+1} = (I + \alpha_n \partial g)^{-1} \big(\tilde{w}^n - \alpha_n \nabla f(\tilde{w}^n) + \beta_n (\tilde{w}^n - \tilde{w}^{n-1})\big)$
		\If {$\|\tilde{w}^{n+1} - \tilde{w}^{n}\|_2 < \varepsilon$}
		\State \textbf{break}
		\EndIf
		\State $n = n + 1$
		\EndWhile
		\State \textbf{return} $\tilde{w}^{n+1}$
	\end{algorithmic}
\end{algorithm}

\section{Numerical studies}\label{sec:num}

In this section, to further investigate the { performance} of our proposed algorithm, we design a series of numerical studies to compare aRSGTSVM and aRSGTSVR with some well-known and recent robust methods to verify the { performance} of our proposed aRSGTSVM and aRSGTSVR.

For the classification problem, we compared the classical 1-SVM \cite{bradley1998feature}, TPMSVM \cite{peng2011tpmsvm}, Pin-TSVM \cite{Xu2017Novel},  rhingeSVM \cite{Xu2017Robust}, and RoBoSS-SVM \cite{Akhtar2025RoBoSS}. The regularization parameter $\lambda$ in 1-SVM is selected from the candidate set $\Lambda = \{0.01, 0.02, 0.03, \ldots, 0.99\}$. For TPMSVM, we select the values of $\nu$ from the set $\{0.1,0.2,\dots,0.8,0.9\}$. For Pin-TSVM, the parameter $\nu$ from set $\{2^{i} \mid i = -8, -7, \dots, 7, 8\}$ and $\tau$ is tuned in $\{0.1,0.3,0.5,0.7,0.9\}$. In the rhingeSVM, the $\eta$ is assumed to be 1. For RoBoSS-SVM, the parameters $a$ and $\lambda$ are selected from the set $\{1,2,3,4,5\}$ and $\{0.5,1,1.5,2\}$ , respectively. For the aRSGTSVM, the parameters $a$ and $\lambda$ are the same as those of RoBoSS-SVM, and we take $\tau$ in the range $\{0.2,0.5,0.8\}$. 

For the regression problem, we compared aRSGTSVR with SVR, LASSO, Elastic Net, TSVR \cite{peng2010tsvr} and Res-TSVR \cite{singla2020robust}. In the SVR, the parameter $\varepsilon$ is optimized in the range $\{2^{i} \mid i = -8, -7, \dots, 7, 8\}$. For TSVR, we set $C_1 = C_2$ and $\varepsilon_{1} = \varepsilon_{2}$. For Res-TSVR, the regularization parameters $C_1 = C_2$ and $\eta$ in the  range of $\{1,3,\dots,16\}$. For the aRSGTSVR, the hyperparameters $a$, $\lambda$, and $\tau$ are optimized through a grid search over the discrete sets $\{1, 2, 3, 4, 5\}$, $\{0.5, 1, 1.5, 2\}$, and $\{0.2, 0.5, 0.8\}$, respectively.

For all of the methods mentioned above, we have chosen the values of $C$ and $\varepsilon$ from the following sets:  $\{2^{i} \mid i = -8, -7, \dots, 7, 8\}$. In the nonlinear case, we consider the Gaussian kernel function, i.e. $ K(x_1, x_2) = \exp(-\gamma \|x_1 - x_2\|_2^2) $. The parameters $\gamma$ of the Gaussian kernel function take values in the range $\{2^{i} \mid i = -8, -7, \dots, 7, 8\}$. { No training-test partition was performed in this study. All models were trained on the entire dataset, and the reported accuracy values are averaged results from five-fold cross-validation, with each fold acting as the validation set sequentially.}

For classification problems, using acc (accuracy) as our evaluation metric. For regression problems, we evaluated the performance of the model using RMSE. MAE was used during the simulations to select the optimal parameters. Their specific definitions are as follows:
\begin{equation}
	\text{RMSE} = \sqrt{\frac{1}{n} \sum_{i=1}^{n} (y_i - \hat{y_i})^2},\quad
	\text{MAE} = \frac{1}{n} \sum_{i=1}^{n} |y_i - \hat{y_i}|,
\end{equation}
where $n$ represents the number of samples used for testing.

\subsection{Artificial datasets for classification}

\begin{figure}[!t]
	\centering
	\subfigure[1-SVM\label{subfig:l1_1}]{
		\includegraphics[width=0.31\textwidth]{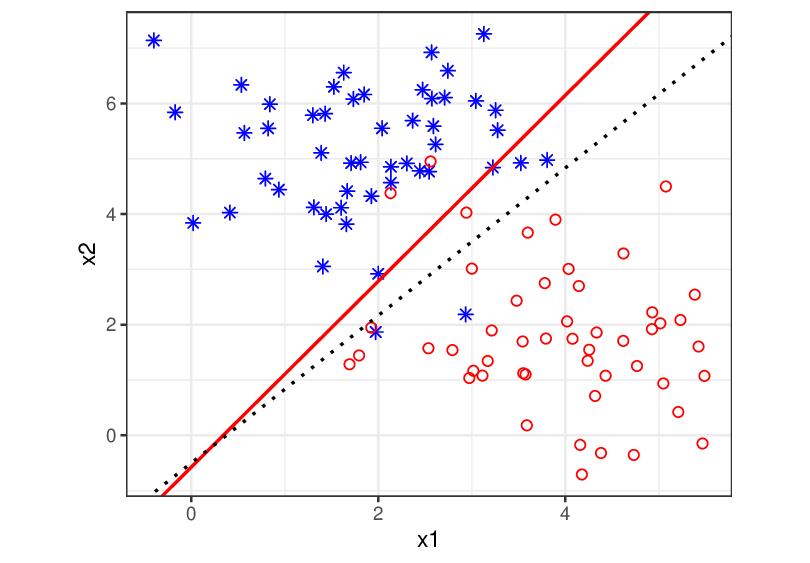}
	}\hfill
	\subfigure[TPMSVM\label{subfig:tpmsvm_1}]{
		\includegraphics[width=0.31\textwidth]{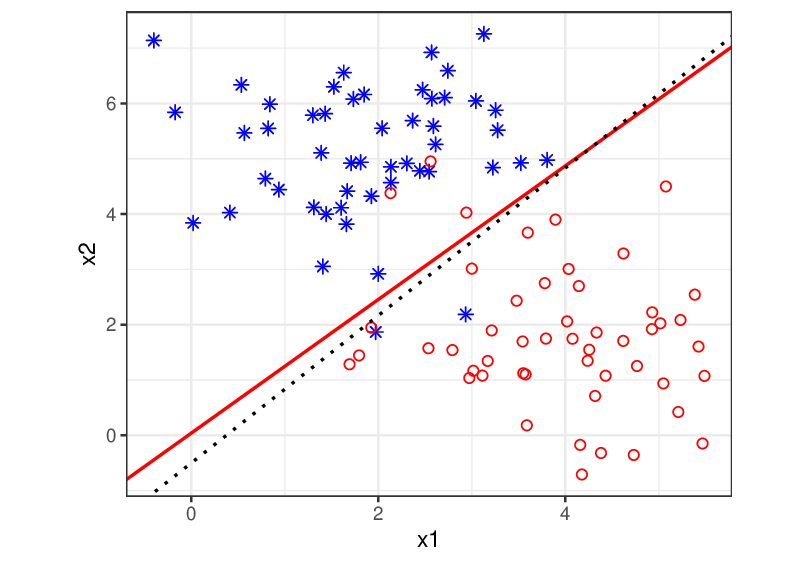}
	}\hfill
	\subfigure[Pin-TSVM\label{subfig:pintsvm_1}]{
		\includegraphics[width=0.31\textwidth]{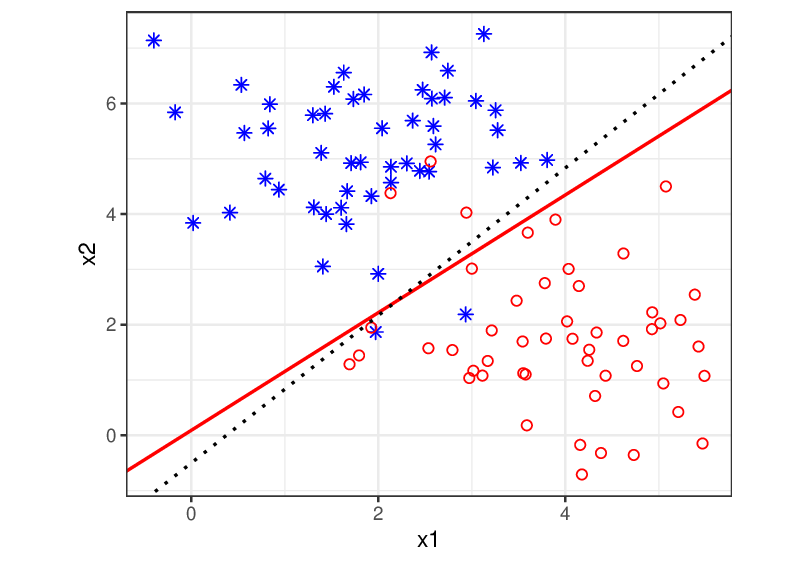}
	}\hfill
	\subfigure[rhingeSVM\label{subfig:rhinge_1}]{
		\includegraphics[width=0.31\textwidth]{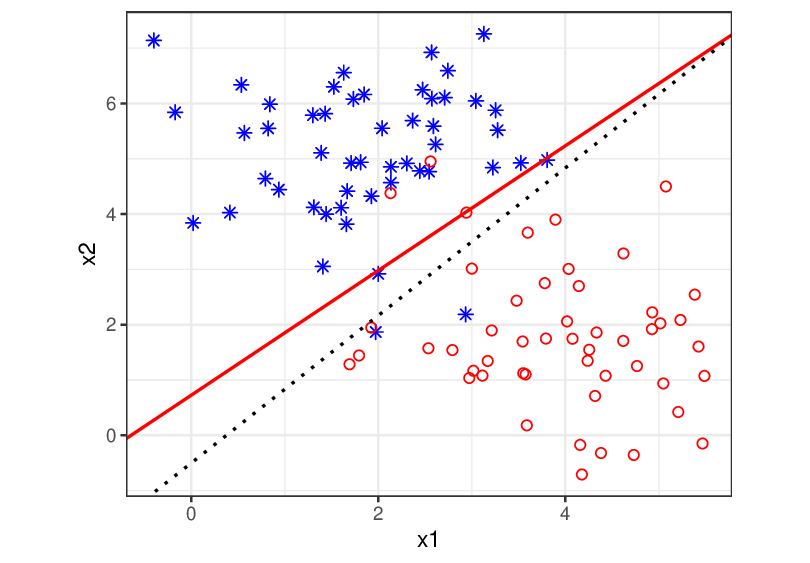}
	}\hfill
	\subfigure[RoBoSS-SVM\label{subfig:roboss_1}]{
		\includegraphics[width=0.31\textwidth]{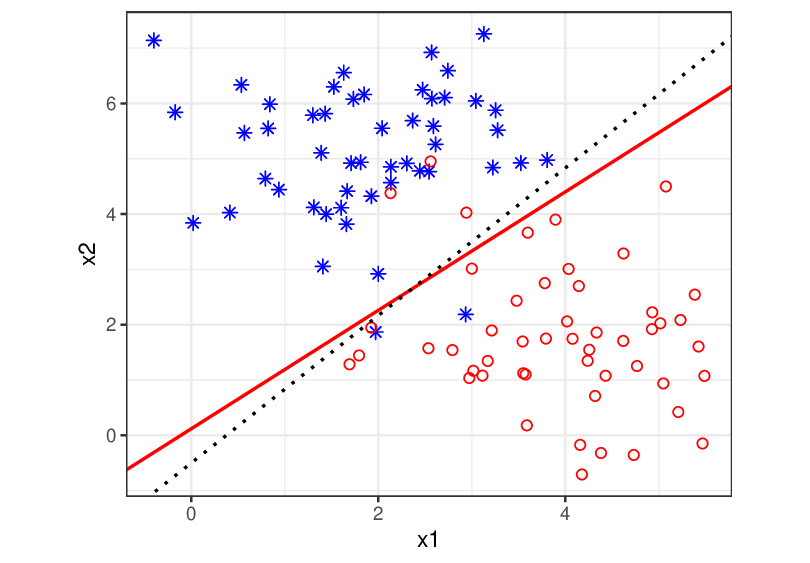}
	}\hfill
	\subfigure[aRSGTSVM\label{subfig:aRSGTSVM_1}]{
		\includegraphics[width=0.31\textwidth]{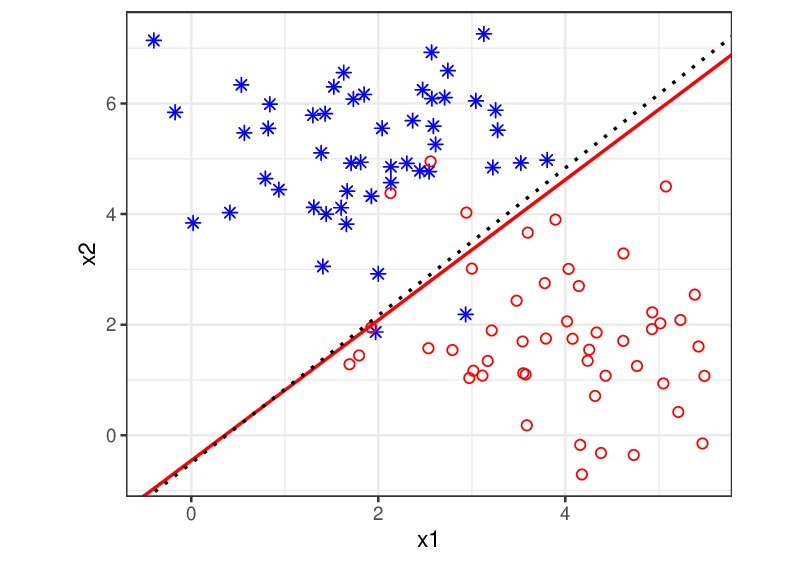}
	}
	\caption{
		The red solid lines represent the separating hyperplanes obtained by 1-SVM, TPMSVM, Pin-TSVM, rhingeSVM, RoBoSS-SVM and aRSGTSVM. The Bayes classifier is shown as a black dotted line.}
	\label{linear 1.1}
\end{figure}

\begin{table}[!t]
	\centering
	\caption{The accuracy(acc) and standard deviation (sd) with linear kernel. The bold is the best one.}
	\label{linear 1.2}
	\resizebox{\textwidth}{!}{
		\begin{tabular}{lcccccc}
			\toprule
			\textbf{label noise} & \textbf{1-SVM} & \textbf{TPMSVM} & \textbf{Pin-TSVM} & \textbf{rhingeSVM} & \textbf{\text{RoBoSS}-\text{SVM}} & \textbf{aRSGTSVM} \\
			& \textbf{acc} $\pm$ \textbf{sd} & \textbf{acc} $\pm$ \textbf{sd} & \textbf{acc} $\pm$ \textbf{sd}  & \textbf{acc} $\pm$ \textbf{sd} & \textbf{acc} $\pm$ \textbf{sd} & \textbf{acc} $\pm$ \textbf{sd}\\ \midrule
			no noise &  0.950 $\pm$ 0.035 & 0.950 $\pm$ 0.055 & 0.950 $\pm$ 0.035 & 0.960 $\pm$ 0.055 & 0.950 $\pm$ 0.035 & \textbf{0.960} $\pm$ \textbf{0.022} \\
			5\% &  0.910 $\pm$ 0.082 & 0.910 $\pm$ 0.004 & 0.920 $\pm$ 0.057 & 0.910 $\pm$ 0.042 & 0.910 $\pm$ 0.065 & \textbf{0.920} $\pm$ \textbf{0.027}
			\\			
			10\% &  0.850 $\pm$ 0.087 & 0.850 $\pm$ 0.004 & 0.850 $\pm$ 0.061 & 0.850 $\pm$ 0.061 & 0.850 $\pm$ 0.094 & \textbf{0.870} $\pm$ \textbf{0.057} \\	
			20\% &  0.780 $\pm$ 0.104 & 0.770 $\pm$ 0.004 & 0.770 $\pm$ 0.130 & 0.770 $\pm$ 0.168 & 0.770 $\pm$ 0.076 & \textbf{0.790} $\pm$ \textbf{0.065} \\
			35\% &  0.620 $\pm$ 0.076 & 0.630 $\pm$ 0.004 & 0.630 $\pm$ 0.091 & 0.630 $\pm$ 0.076 & 0.630 $\pm$ 0.065 & \textbf{0.660} $\pm$ \textbf{0.074} \\
			\bottomrule
		\end{tabular}
	}
\end{table}

\textbf{Example 1.} To evidence the robustness of aRSGTSVM to label noise, we conducted an experiment using a two-dimensional synthetic dataset containing 100 samples. In this example, the samples were generated from two Gaussian distributions with equal probability:
$x_i, i \in \{1, 2, \ldots, 50\} \sim N(u_1, \Sigma_1), x_i, i \in \{51, 52, \ldots, 100\} \sim N(u_2, \Sigma_2)$, where $u_1 = [2, 5]^T, u_2 = [4, 2]^T$ and $\Sigma_1 = \Sigma_2 = \operatorname{diag}(1, 2)$. In this case, the Bayes classifier is $f_C(x) = 4x_1 - 3x_2 - \frac{3}{2}$. To systematically investigate the impact of label noise on model performance, experiments were conducted under varying noise ratios.

In Figure \ref{linear 1.1} presents a noise-free condition, in which the black dotted line represents the Bayesian optimal decision boundary and the red solid line represents the decision boundary obtained from each model. Among the six models evaluated, our proposed aRSGTSVM demonstrates the most satisfactory performance, with its decision boundary showing the closest alignment to that of the Bayesian classifier. The complete experimental results are presented in Table \ref{linear 1.2}, which demonstrate the remarkable robustness and noise insensitivity of our proposed aRSGTSVM.

\begin{figure}[!t]
	\centering
	\subfigure[1-SVM\label{subfig:l1_2}]{
		\includegraphics[width=0.31\textwidth]{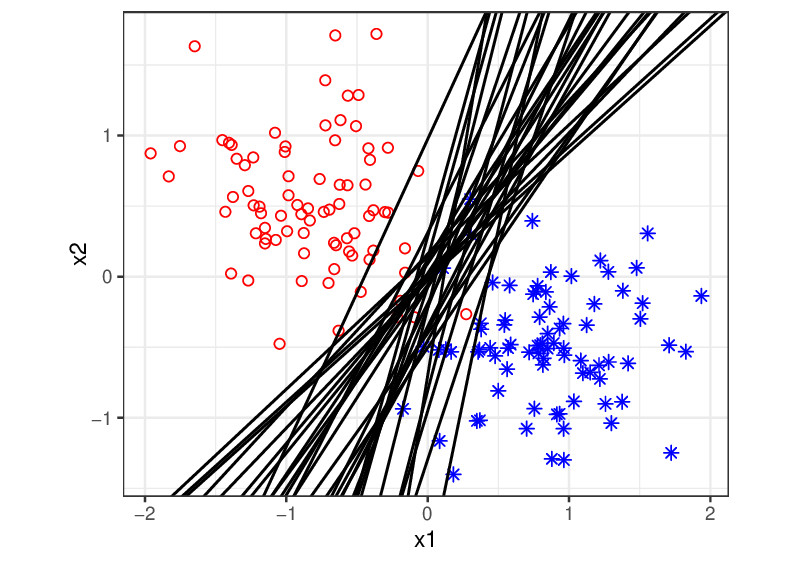}
	}\hfill
	\subfigure[TPMSVM\label{subfig:tpmsvm_2}]{
		\includegraphics[width=0.31\textwidth]{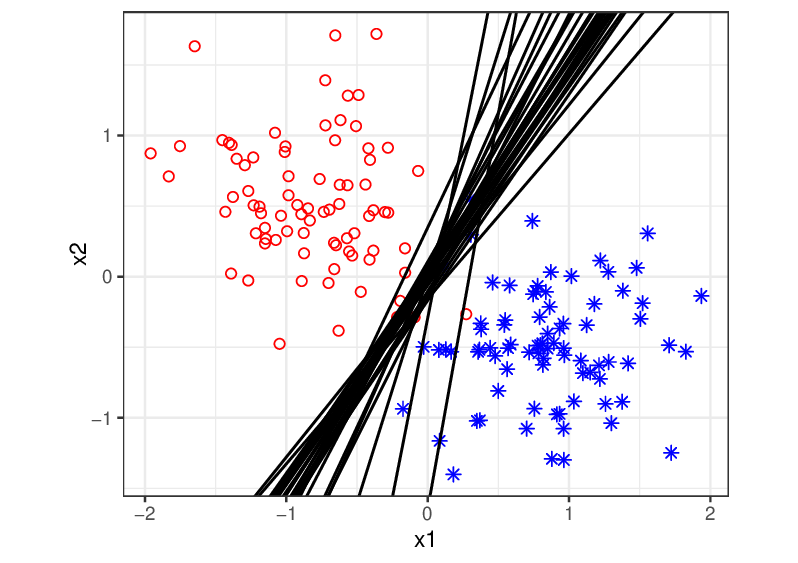}
	}\hfill
	\subfigure[Pin-TSVM\label{subfig:pintsvm_2}]{
		\includegraphics[width=0.31\textwidth]{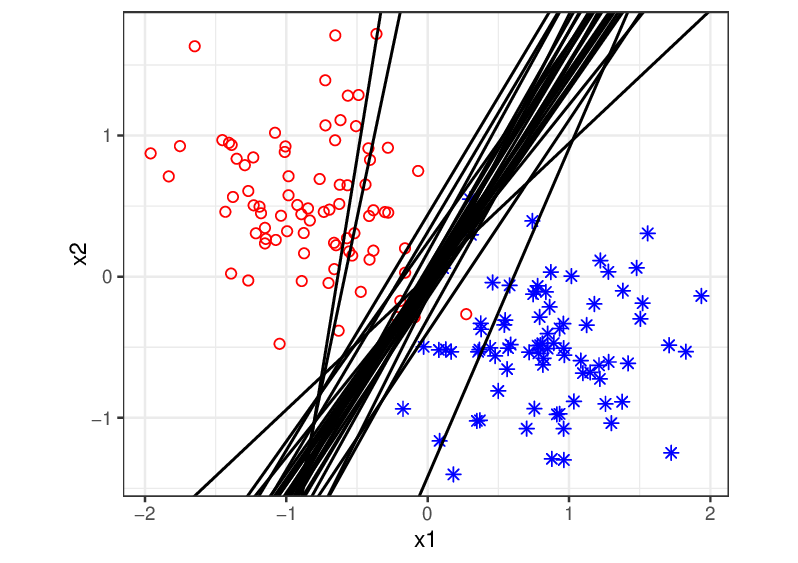}
	}\hfill
	\subfigure[rhingeSVM\label{subfig:rhinge_2}]{
		\includegraphics[width=0.31\textwidth]{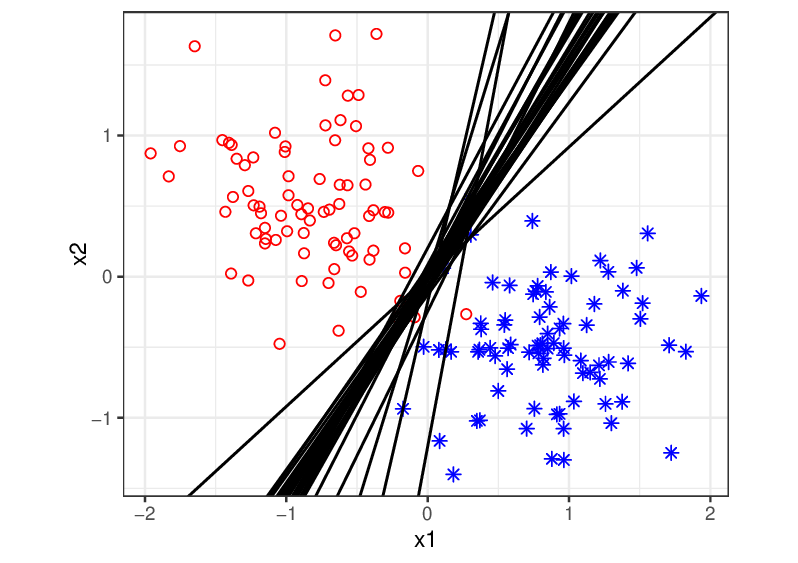}
	}\hfill
	\subfigure[RoBoSS-SVM\label{subfig:roboss_2}]{
		\includegraphics[width=0.31\textwidth]{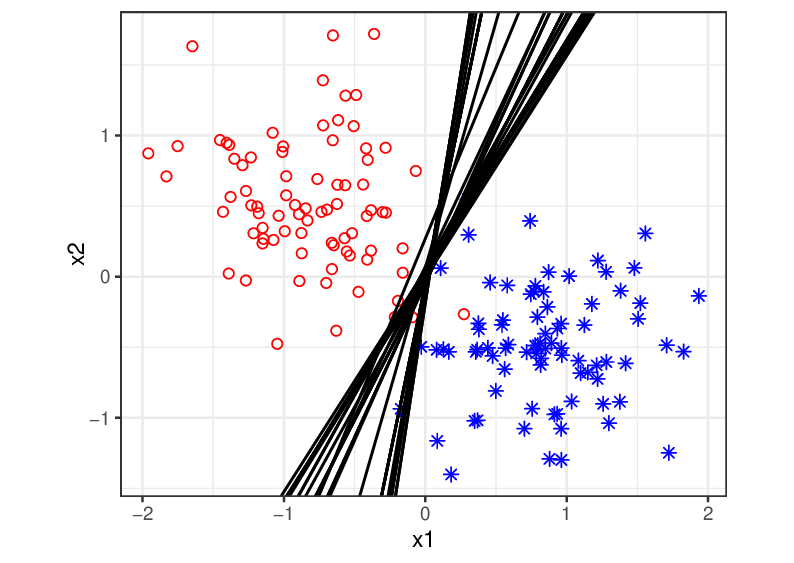}
	}\hfill
	\subfigure[aRSGTSVM\label{subfig:aRSGTSVM_2}]{
		\includegraphics[width=0.31\textwidth]{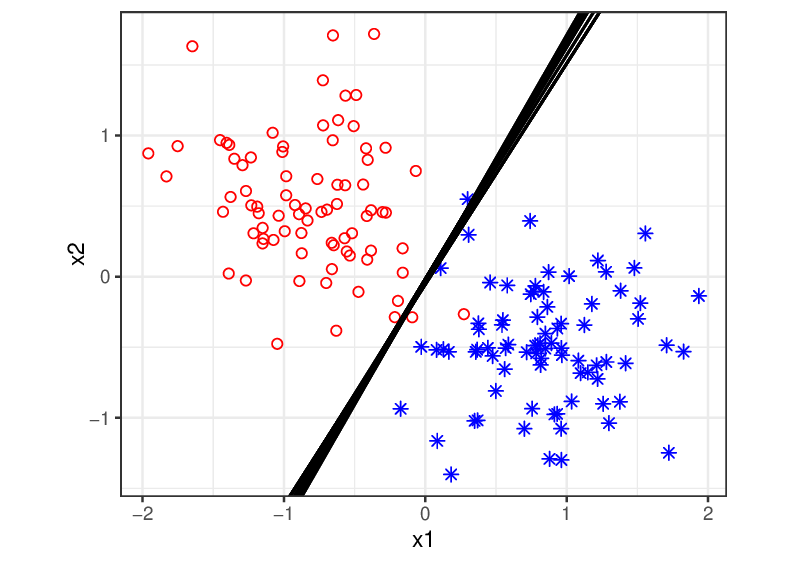}
	}
	\caption{
		The black solid lines represent the separating hyperplanes obtained by 1-SVM, TPMSVM, Pin-TSVM, rhingeSVM, RoBoSS-SVM and aRSGTSVM.}
	\label{linear 2}
\end{figure}

\textbf{Example 2.} To exemplify the stability of SVM under resampling, a total of 160 samples were generated based on Example 1, with the parameters set to $u_1 = [0.8, -0.5]^T$, $u_2 = [-0.8, 0.5]^T$, and $\Sigma_1 = \Sigma_2 = \operatorname{diag}(0.2, 0.2)$. The experiment was repeated 30 times.

\begin{table}[!t]
	\centering
	\caption{The accuracy (acc) and standard deviation (sd) in different $(n, p)$. The bold is the best one.}
	\label{linear 3}
	\resizebox{\textwidth}{!}{
		\begin{tabular}{lllllll}
			\toprule
			\multicolumn{7}{c}{(a) 0\% label noise} \\
			\midrule
			\textbf{} & \textbf{1-SVM} & \textbf{TPMSVM} & \textbf{Pin-TSVM} & \textbf{rhingeSVM} & \textbf{\text{RoBoSS}-\text{SVM}} & \textbf{aRSGTSVM} \\
			\midrule
			$n=50,p=100$ & 0.900 $\pm$ 0.071 & 0.920 $\pm$ 0.130 & \textbf{0.940} $\pm$ \textbf{0.055} & 0.900 $\pm$ 0.071 & 0.920 $\pm$ 0.110 & 0.940 $\pm$ 0.089
			\\	
			$n=50,p=150$ & 0.940 $\pm$ 0.055 & 0.920 $\pm$ 0.084 & 0.940 $\pm$ 0.055 & 0.920 $\pm$ 0.084 & 0.920 $\pm$ 0.084 & \textbf{0.960} $\pm$ \textbf{0.055}
			\\	
			$n=100,p=150$ & 0.920 $\pm$ 0.027 & 0.930 $\pm$ 0.027 & 0.930 $\pm$ 0.076 & 0.920 $\pm$ 0.067 & 0.930 $\pm$ 0.057 & \textbf{0.970} $\pm$ \textbf{0.045}
			\\	
			$n=100,p=200$ & 0.920 $\pm$ 0.076 & 0.910 $\pm$ 0.055 & 0.920 $\pm$ 0.084 & 0.910 $\pm$ 0.074 & 0.930 $\pm$ 0.045 & \textbf{0.960} $\pm$ \textbf{0.042}
			\\	
			\midrule
			\multicolumn{7}{c}{(b) 5\% label noise} \\
			\midrule
			$n=50,p=100$ & 0.860 $\pm$ 0.114 & 0.880 $\pm$ 0.084 & 0.900 $\pm$ 0.122 & 0.860 $\pm$ 0.134 & 0.880 $\pm$ 0.045 & \textbf{0.920} $\pm$ \textbf{0.084}
			\\	
			$n=50,p=150$ & 0.940 $\pm$ 0.055 & 0.900 $\pm$ 0.071 & 0.940 $\pm$ 0.055 & 0.920 $\pm$ 0.045 & 0.940 $\pm$ 0.089 & \textbf{0.960} $\pm$ \textbf{0.055}
			\\	
			$n=100,p=150$ & 0.900 $\pm$ 0.035 & 0.880 $\pm$ 0.057 & 0.890 $\pm$ 0.108 & 0.880 $\pm$ 0.084 & 0.900 $\pm$ 0.100 & \textbf{0.940} $\pm$ \textbf{0.042}
			\\	
			$n=100,p=200$ & 0.930 $\pm$ 0.084 & 0.920 $\pm$ 0.076 & 0.920 $\pm$ 0.057 & 0.910 $\pm$ 0.042 & 0.930 $\pm$ 0.045 & \textbf{0.950} $\pm$ \textbf{0.035}
			\\	
			\midrule
			\multicolumn{7}{c}{(c) 10\% label noise} \\
			\midrule
			$n=50,p=100$ & 0.840 $\pm$ 0.134 & 0.840 $\pm$ 0.152 & 0.860 $\pm$ 0.114 & 0.840 $\pm$ 0.114 & 0.860 $\pm$ 0.055 & \textbf{0.880} $\pm$ \textbf{0.130}
			\\	
			$n=50,p=150$ & 0.900 $\pm$ 0.100 & 0.880 $\pm$ 0.084 & 0.880 $\pm$ 0.084 & 0.880 $\pm$ 0.164 & 0.900 $\pm$ 0.071 & \textbf{0.920} $\pm$ \textbf{0.084}
			\\	
			$n=100,p=150$ & 0.830 $\pm$ 0.104 & 0.840 $\pm$ 0.114 & 0.830 $\pm$ 0.076 & 0.810 $\pm$ 0.108 & 0.850 $\pm$ 0.087 & \textbf{0.870} $\pm$ \textbf{0.097}
			\\	
			$n=100,p=200$ & 0.830 $\pm$ 0.097 & 0.800 $\pm$ 0.079 & 0.810 $\pm$ 0.082 & 0.810 $\pm$ 0.124 & 0.830 $\pm$ 0.144 & \textbf{0.870} $\pm$ \textbf{0.027}
			\\	
			\midrule
			\multicolumn{7}{c}{(d) 20\% label noise} \\
			\midrule
			$n=50,p=100$ & 0.720 $\pm$ 0.228 & 0.700 $\pm$ 0.122 & 0.740 $\pm$ 0.207 & 0.720 $\pm$ 0.130 & 0.720 $\pm$ 0.179 & \textbf{0.780} $\pm$ \textbf{0.110}
			\\	
			$n=50,p=150$ & 0.800 $\pm$ 0.071 & 0.840 $\pm$ 0.089 & 0.780 $\pm$ 0.110 & 0.780 $\pm$ 0.110 & 0.780 $\pm$ 0.084 & \textbf{0.840} $\pm$ \textbf{0.195}
			\\	
			$n=100,p=150$ & 0.770 $\pm$ 0.130 & 0.770 $\pm$ 0.057 & 0.750 $\pm$ 0.112 & 0.770 $\pm$ 0.115 & 0.770 $\pm$ 0.084 & \textbf{0.800} $\pm$ \textbf{0.087}
			\\	
			$n=100,p=200$ & 0.770 $\pm$ 0.115 & 0.760 $\pm$ 0.134 & 0.760 $\pm$ 0.055 & 0.740 $\pm$ 0.089 & 0.770 $\pm$ 0.104 & \textbf{0.790} $\pm$ \textbf{0.089}
			\\	
			\midrule
			\multicolumn{7}{c}{(e) 35\% label noise} \\
			\midrule
			$n=50,p=100$ & 0.600 $\pm$ 0.100 & 0.600 $\pm$ 0.141 & 0.620 $\pm$ 0.084 & 0.560 $\pm$ 0.182 & 0.660 $\pm$ 0.195 & \textbf{0.680} $\pm$ \textbf{0.130}
			\\	
			$n=50,p=150$ & 0.700 $\pm$ 0.071 & 0.660 $\pm$ 0.167 & 0.740 $\pm$ 0.089 & 0.660 $\pm$ 0.152 & 0.720 $\pm$ 0.148 & \textbf{0.780} $\pm$ \textbf{0.205}
			\\	
			$n=100,p=150$ & 0.620 $\pm$ 0.057 & 0.610 $\pm$ 0.108 & 0.630 $\pm$ 0.097 & 0.600 $\pm$ 0.127 & 0.640 $\pm$ 0.089 & \textbf{0.660} $\pm$ \textbf{0.082}
			\\	
			$n=100,p=200$ & 0.620 $\pm$ 0.115 & 0.600 $\pm$ 0.035 & 0.640 $\pm$ 0.055 & 0.590 $\pm$ 0.152 & 0.620 $\pm$ 0.104 & \textbf{0.670} $\pm$ \textbf{0.067}
			\\				
			\bottomrule
		\end{tabular}
	}
\end{table}

{ As visualized in Figure \ref{linear 2}, the proposed aRSGTSVM demonstrates exceptional robustness and stability in constructing decision hyperplanes. Specifically, the hyperplanes of aRSGTSVM form a remarkably narrow and highly concentrated cluster, with almost no variation across different resampling runs, highlighting its strong resistance to sample perturbations. While RoBoSS-SVM exhibits improved stability relative to traditional SVM variants, its hyperplanes still show non-negligible dispersion. In sharp contrast, the baseline models, including 1-SVM, TPMSVM, Pin-TSVM and rhingeSVM, generate widely scattered hyperplanes with substantial fluctuations, particularly in the boundary regions between classes. This dispersion reveals their sensitivity to minor changes in the training data, resulting in unstable decision boundaries. Beyond its robustness advantage, aRSGTSVM also achieves the most accurate separation between the two classes (red and blue points), effectively distinguishing between the samples and delivering the best overall classification performance among all evaluated models.}

\textbf{Example 3.} In this example, the samples were generated from two multivariate Gaussian distributions with equal probability:
$x_i, i \in \{1, 2, \ldots, \dfrac{n}{2}\} \sim N(u_1, \Sigma_1), x_i, i \in \{\dfrac{n}{2}+1, \dfrac{n}{2}+2, \ldots, n\} \sim N(u_2, \Sigma_2)$, where $u_1 = (0.1, 0.2, 0.3, 0.4, 0.5, 0, \ldots, 0)^T \in \mathbb{R}^{n}$, $u_2 = -u_1$ and $\Sigma_1 = \Sigma_2 = (\sigma_{ij})$ is defined with non-zero elements:
$\sigma_{ii} = 1, \text{for } i = 1, 2, \ldots, n$ and $\sigma_{ij} = -0.2, \text{for } 1 \leq i \neq j \leq 5$.

{ As shown in Table \ref{linear 3}, we systematically compared the classification performance of the proposed aRSGTSVM with several state-of-the-art SVM-based methods across different combinations of sample size $n$, feature dimension $p$, and varying label noise levels ranging from 0\% to 35\%. The results, measured by mean classification accuracy and standard deviation, consistently demonstrate that aRSGTSVM achieves the highest or tied-for-highest accuracy in nearly all experimental settings, including high-dimensional scenarios where $p > n$, while maintaining stable performance across repeated trials. As label noise increases, aRSGTSVM consistently outperforms competing methods, showing its superiority in both clean high-dimensional environments and noisy conditions. This comprehensive comparison confirms that the proposed aRSGTSVM exhibits stronger robustness against label noise and better adaptability to high-dimensional data compared with baseline models, validating the effectiveness of its asymmetric regularized twin SVM framework.}

\begin{figure}[!t]
    \centering
    \includegraphics[width=4.5in]{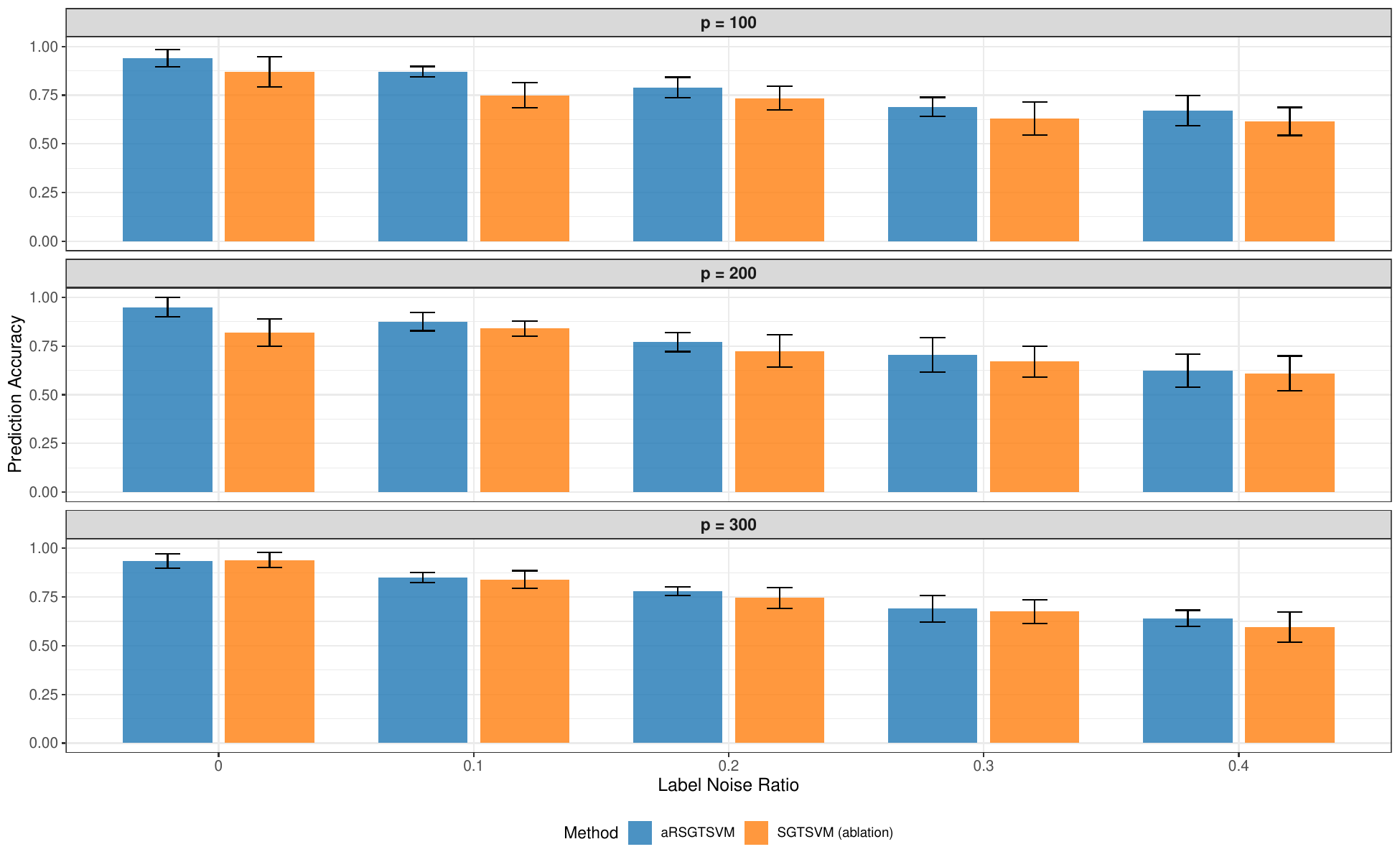}
    \caption{The results of the ablation experiments under different ratios of label noise and number of features.}
    \label{fig:ablation}
\end{figure}

{
\textbf{Example 4.} To further verify the effectiveness of the proposed asymmetric RoBoSS loss function, we conduct ablation experiments. Based on the setting of Example 3, we fix the sample size at \(n=200\), set the feature dimension \(p \in \{100, 200, 300\}\) and the label noise ratio to \(\{0\%, 10\%, 20\%, 30\%, 40\%\}\). The experimental results are presented in Figure \ref{fig:ablation}.

As shown in Figure \ref{fig:ablation}, under different feature dimensions ($p = 100, 200, 300$) and varying label noise ratios ($0\%$ to $40\%$), the proposed aRSGTSVM consistently achieves higher prediction accuracy than the ablation method SGTSVM, which removes the asymmetric robust loss. As the noise ratio increases from $0\%$ to $40\%$, the prediction accuracy of both methods decreases. However, aRSGTSVM exhibits a much smaller decline, demonstrating superior robustness to label noise. Moreover, as the feature dimension increases from $100$ to $300$, aRSGTSVM maintains stable performance, whereas SGTSVM shows a more pronounced drop in accuracy. These results validate the effectiveness and robustness of the proposed asymmetric RoBoSS loss function in scenarios with label noise and high-dimensional features.
}

\begin{figure}[!t]
    \centering
    \includegraphics[width=4.5in]{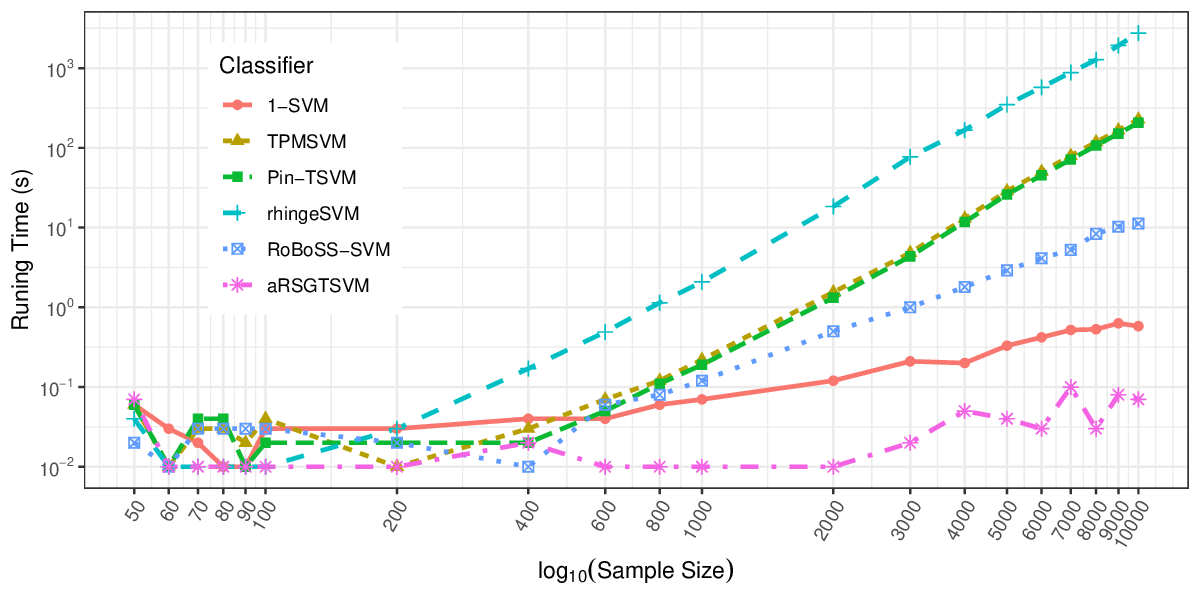}
    \caption{The one-run CPU time cost of 1-SVM, TPMSVM, Pin-TSVM, rhingeSVM, RoBoSS-SVM and aRSGTSVM under different sample sizes. }
    \label{fig:time_n}
\end{figure}

\begin{figure}[!t]
    \centering
    \includegraphics[width=4.5in]{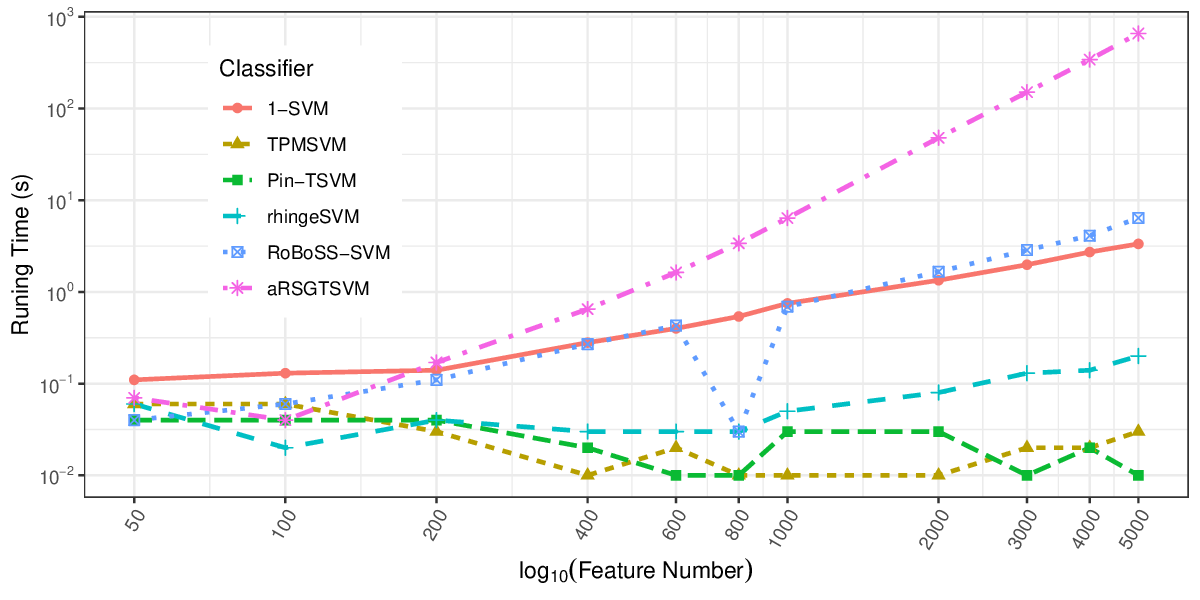}
    \caption{The one-run CPU time cost of 1-SVM, TPMSVM, Pin-TSVM, rhingeSVM, RoBoSS-SVM and aRSGTSVM under different feature numbers. }
    \label{fig:time_p}
\end{figure}

{
\textbf{Example 5.} In this example, we investigate the CPU time cost of our proposed algorithm (Algorithm~\ref{algorithm1}) with varying numbers of samples and features, respectively. Based on the setting of Example 1, we fix the feature dimension at \(p=2\), set the sample size n ranging from 50 to 10000 with a label noise ratio of \(15\%\), and set all parameters to 1 except for \(\tau=0.5\). The results are presented in Figure \ref{fig:time_n}. Based on the setting of Example 3, We further conduct experiments with a fixed sample size \(n=100\) and feature dimension p ranging from 50 to 5000, under the same noise ratio and parameter settings, as shown in Figure \ref{fig:time_p}.

Figure \ref{fig:time_n} presents the CPU running time of various classifiers under different sample sizes. As the sample size increases from 50 to 10000, the running time of most compared algorithms increases significantly. Among them, rhingeSVM and Pin-TSVM show the fastest growth in time complexity, with their time consumption far exceeding other models in large-sample scenarios. In contrast, the proposed aRSGTSVM maintains an extremely low running time throughout. Even when the sample size reaches 10000, its time consumption remains at a low level, only slightly higher than 1-SVM and much lower than other compared models, demonstrating excellent scalability to large sample sizes.

Figure \ref{fig:time_p} illustrates the single-run CPU time (log scale) of different algorithms as the feature dimension increases from 50 to 5,000, with a fixed sample size of $n = 100$. For the proposed aRSGTSVM, the running time grows slowly when the number of features is below 200, but increases sharply thereafter, becoming the highest among all compared methods at 5,000 dimensions. In contrast, Pin-TSVM and TPMSVM maintain consistently low running times, while 1-SVM and RoBoSS-SVM exhibit an approximately linear increase with a much lower growth rate than aRSGTSVM. In summary, aRSGTSVM is sensitive to feature dimensionality, incurring high computational costs in high-dimensional scenarios, though its running time remains acceptable in low-to-medium dimensions.
}

\subsection{Artificial datasets for regression}

\textbf{Example 1.} In this example, we designed a two-dimensional case to demonstrate that aRSGTSVR is robust under different types of noise, namely uniform noise and Gaussian noise. We generate 160 samples as follows:
\begin{equation}
	y = \operatorname{sinc}(x_i) = \frac{\sin x_i}{x_i} + e_i, \quad x_i \sim U[-3\pi, 3\pi].
	\label{sinc}
\end{equation}

Noise is added based on  (\ref{sinc}), which is defined as follows:
\begin{equation*}
	\text{Type A: } e_i \sim U[-0.4, 0.4],  \text{Type B: } e_i \sim N(0, 0.2^2).
\end{equation*}

\begin{figure}[!t]
	\centering
	\subfigure[No noise]{
		\includegraphics[width=0.31\textwidth]{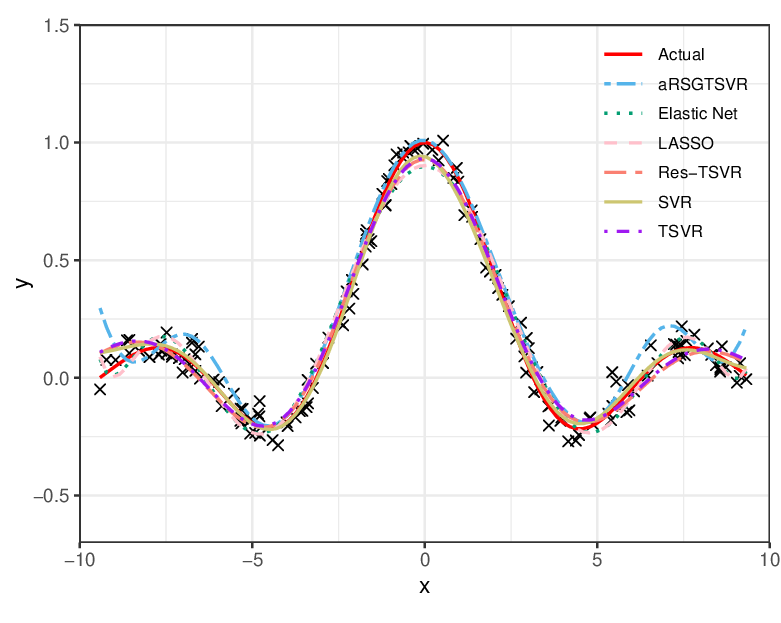}
	}
	\hfill
	\subfigure[Uniform noise]{
		\includegraphics[width=0.31\textwidth]{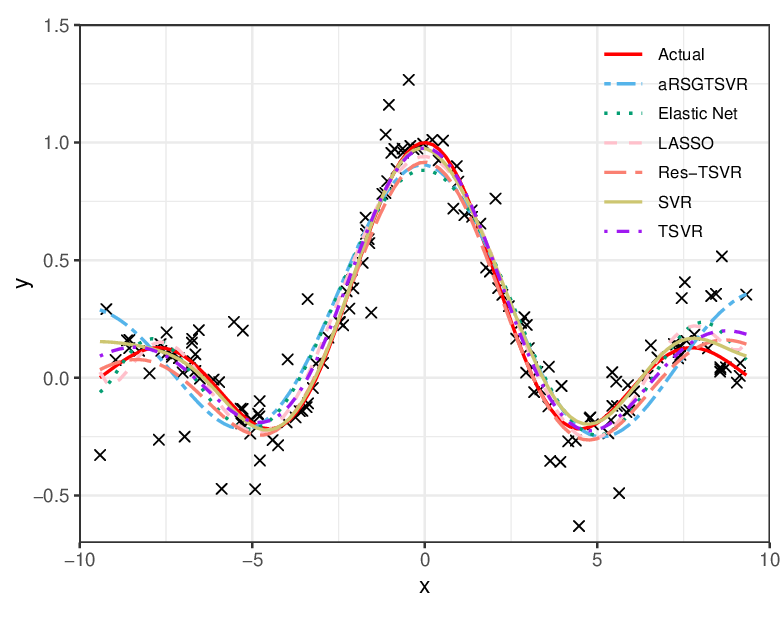}
	}
	\hfill
	\subfigure[Gaussian noise]{
		\includegraphics[width=0.31\textwidth]{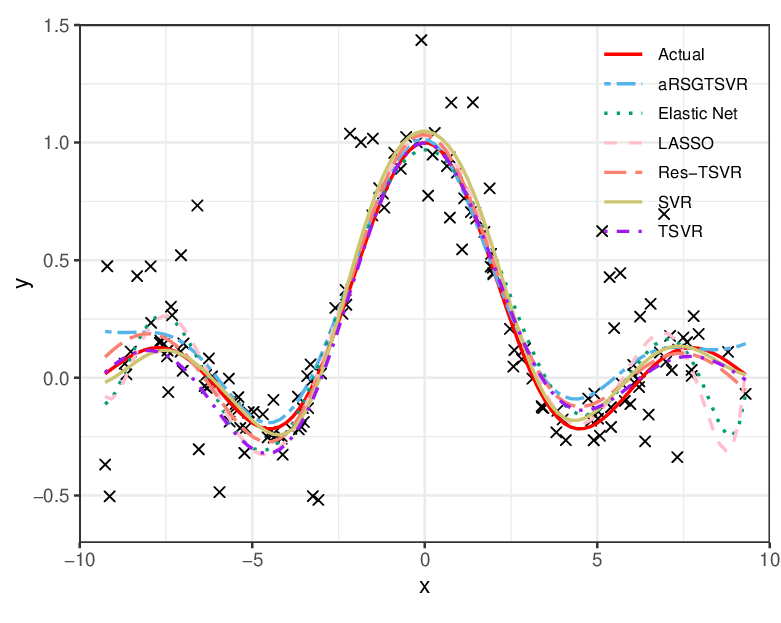}
	}
	\caption{Predictions of SVR, TSVR, Res-TSVR and aRSGTSVR on Sinc function with different noises.}
	\label{nonlinear 1.1}
\end{figure}

We plotted the fitting curve of each model in Figure \ref{nonlinear 1.1}. Since it is not intuitively obvious which model fits the true curve better, we present the corresponding root mean square error and standard deviation { obtained with the nonlinear kernel setting in Table \ref{nonlinear 1.2}. The results consistently show that the proposed aRSGTSVR achieves the best overall performance in terms of RMSE across different noise conditions, including noiseless, uniform noise, and Gaussian noise scenarios, while maintaining a relatively low standard deviation. Compared with competing methods such as SVR, LASSO, Elastic Net, TSVR, and Res-TSVR, aRSGTSVR exhibits more stable and superior regression performance, which confirms that our method is more robust than other comparative models under different types of noise when handling nonlinear regression tasks.}

\begin{table}[!t]
	\centering
	\scriptsize
	\caption{The RMSE and standard deviation (sd) with nonlinear kernel. The bold is the best one.}
	\label{nonlinear 1.2}
	\resizebox{\textwidth}{!}{%
		\begin{tabular}{lllllll}
			\toprule
			\textbf{} & \textbf{SVR} & \textbf{LASSO} & \textbf{Elastic Net} & \textbf{TSVR} & \textbf{Res-TSVR} & \textbf{aRSGTSVR} \\
			\midrule
			No noise & 0.080 $\pm$ 0.013 & 0.074 $\pm$ 0.012 & 0.075 $\pm$ 0.011 & 0.075 $\pm$ 0.030 & 0.126 $\pm$ 0.115 & \textbf{0.071 $\pm$ 0.010} \\
			Uniform noise & 0.128 $\pm$ 0.026 & 0.129 $\pm$ 0.014 & 0.130 $\pm$ 0.015 & 0.129 $\pm$ 0.021 & 0.128 $\pm$ 0.022 & \textbf{0.118 $\pm$ 0.017} \\
			Gaussian noise & 0.267 $\pm$ 0.059 & 0.270 $\pm$ 0.062 & 0.265 $\pm$ 0.060 & 0.265 $\pm$ 0.055 & 0.264 $\pm$ 0.057 & \textbf{0.260 $\pm$ 0.073} \\
			\bottomrule
		\end{tabular}%
	}
\end{table}

\begin{table}[!t]
	\centering
	\caption{The RMSE and standard deviation (sd) in different $(n, p)$. The bold is the best one.}
	\label{nonlinear 2.1}
	\resizebox{\textwidth}{!}{
		\begin{tabular}{lllllll}
			\toprule
			\multicolumn{7}{c}{(a) 0\% label noise} \\
			\midrule
			\textbf{} & \textbf{SVR} & \textbf{LASSO} & \textbf{Elastic Net} & \textbf{TSVR} & \textbf{Res-TSVR} & \textbf{aRSGTSVR} \\
			\midrule
			$n=50,p=30$ & 1.570 $\pm$ 1.342 & 1.968 $\pm$ 1.988 & 1.819 $\pm$ 1.838 & 1.949 $\pm$ 1.254 & 1.861 $\pm$ 0.892 & \textbf{1.438} $\pm$ \textbf{0.675}
			\\	
			$n=50,p=50$ & 1.712 $\pm$ 1.902 & 2.331 $\pm$ 2.354 & 2.331 $\pm$ 2.354 & 2.105 $\pm$ 1.998 & 1.976 $\pm$ 0.907 & \textbf{1.481} $\pm$ \textbf{1.041}
			\\	
			$n=50,p=100$ & 1.811 $\pm$ 1.900 & 2.331 $\pm$ 2.354 & 2.331 $\pm$ 2.354 & 2.140 $\pm$ 0.884 & 2.041 $\pm$ 2.532 & \textbf{1.501} $\pm$ \textbf{1.135}
			\\	
			\midrule
			\multicolumn{7}{c}{(b) 5\% label noise} \\
			\midrule
			$n=50,p=30$ & 1.574 $\pm$ 1.341 & 1.958 $\pm$ 1.977 & 1.958 $\pm$ 1.977 & 1.942 $\pm$ 1.469 & 1.862 $\pm$ 0.770 & \textbf{1.406} $\pm$ \textbf{1.101}
			\\	
			$n=50,p=50$ & 1.802 $\pm$ 1.607 & 2.325 $\pm$ 2.349 & 2.325 $\pm$ 2.349 & 2.097 $\pm$ 1.618 & 1.904 $\pm$ 1.142 & \textbf{1.461} $\pm$ \textbf{1.543}
			\\	
			$n=50,p=100$ & 1.773 $\pm$ 1.478 & 2.325 $\pm$ 2.349 & 2.325 $\pm$ 2.349 & 2.157 $\pm$ 2.204 & 1.986 $\pm$ 1.350 & \textbf{1.480} $\pm$ \textbf{1.438}
			\\	
			\midrule
			\multicolumn{7}{c}{(c) 10\% label noise} \\
			\midrule
			$n=50,p=30$ & 1.632 $\pm$ 0.574 & 1.990 $\pm$ 2.010 & 1.990 $\pm$ 2.010 & 1.978 $\pm$ 1.155 & 1.892 $\pm$ 0.922 & \textbf{1.444} $\pm$ \textbf{0.717}
			\\	
			$n=50,p=50$ & 1.835 $\pm$ 0.774 & 2.334 $\pm$ 2.358 & 2.334 $\pm$ 2.358 & 2.178 $\pm$ 2.055 & 2.091 $\pm$ 0.784 & \textbf{1.487} $\pm$ \textbf{1.290}
			\\	
			$n=50,p=100$ & 1.878 $\pm$ 1.115 & 2.334 $\pm$ 2.358 & 2.334 $\pm$ 2.358 & 2.165 $\pm$ 1.645 & 2.120 $\pm$ 1.348 & \textbf{1.483} $\pm$ \textbf{1.279}
			\\	
			\midrule
			\multicolumn{7}{c}{(d) 20\% label noise} \\
			\midrule
			$n=50,p=30$ & 1.590 $\pm$ 0.477 & 1.962 $\pm$ 1.982 & 1.825 $\pm$ 1.843 & 1.946 $\pm$ 1.596 & 1.709 $\pm$ 1.477 & \textbf{1.435} $\pm$ \textbf{0.994}
			\\	
			$n=50,p=50$ & 1.815 $\pm$ 1.876 & 2.336 $\pm$ 2.360 & 2.336 $\pm$ 2.360 & 2.150 $\pm$ 1.257 & 1.941 $\pm$ 1.999 & \textbf{1.485} $\pm$ \textbf{1.315}
			\\	
			$n=50,p=100$ & 1.832 $\pm$ 2.459 & 2.336 $\pm$ 2.360 & 2.336 $\pm$ 2.360 & 2.162 $\pm$ 1.702 & 1.924 $\pm$ 1.491 & \textbf{1.535} $\pm$ \textbf{1.984}
			\\	
			\midrule
			\multicolumn{7}{c}{(e) 35\% label noise} \\
			\midrule
			$n=50,p=30$ & 1.590 $\pm$ 0.360 & 1.972 $\pm$ 1.992 & 1.823 $\pm$ 1.841 & 1.896 $\pm$ 1.044 & 1.858 $\pm$ 1.041 & \textbf{1.400} $\pm$ \textbf{0.959}
			\\	
			$n=50,p=50$ & 1.767 $\pm$ 1.566 & 2.352 $\pm$ 2.376 & 2.352 $\pm$ 2.376 & 2.169 $\pm$ 1.798 & 1.939 $\pm$ 1.758 & \textbf{1.528} $\pm$ \textbf{1.302}
			\\	
			$n=50,p=100$ & 1.787 $\pm$ 1.583 & 2.352 $\pm$ 2.376 & 2.352 $\pm$ 2.376 & 2.157 $\pm$ 2.144 & 2.002 $\pm$ 1.248 & \textbf{1.519} $\pm$ \textbf{1.682}
			\\					
			\bottomrule
		\end{tabular}
	}
\end{table}

\textbf{Example 2.} To illustrate that aRSGTSVR exhibits excellent variable selection capability under different noise ratios for high-dimensional data, in this example, the samples are generated from the following multivariate Gaussian distribution:
$x_i, i \in \{1, 2, \ldots, n\} \sim N(u, \Sigma)$, where $u = (0.1, 0.2, 0.3, 0.4, 0.5, 0, \ldots, 0)^T \in \mathbb{R}^{n}$ and $\Sigma = (\sigma_{ij})$ is defined with non-zero elements:
$\sigma_{ii} = 1, \text{for } i = 1, 2, \ldots, n$ and $\sigma_{ij} = -0.2, \text{for } 1 \leq i \neq j \leq 5$.

The response variable $y$ for each sample is constructed via the function:
\begin{equation}
	y = \sin(3x_1) + x_2^2 + \exp(-|x_3|) + \log(|x_4| + 1) + x_1x_5.
\end{equation}

\begin{figure}[!t]
	\centering
	\subfigure[$n=50,p=30$]{
		\includegraphics[width=0.31\textwidth]{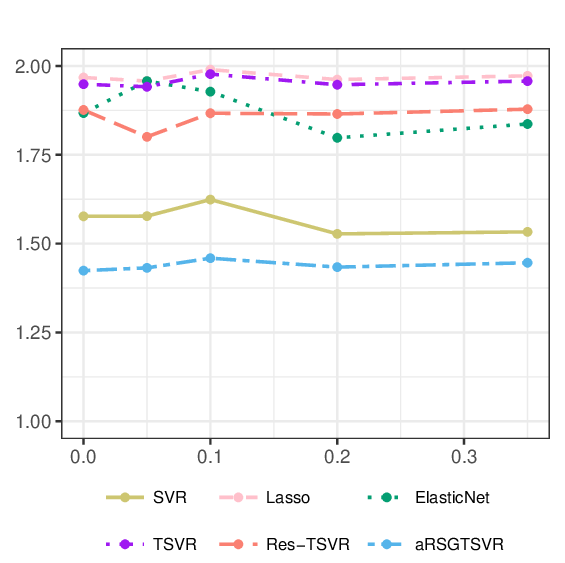}
	}
	\hfill
	\subfigure[$n=50,p=50$]{
		\includegraphics[width=0.31\textwidth]{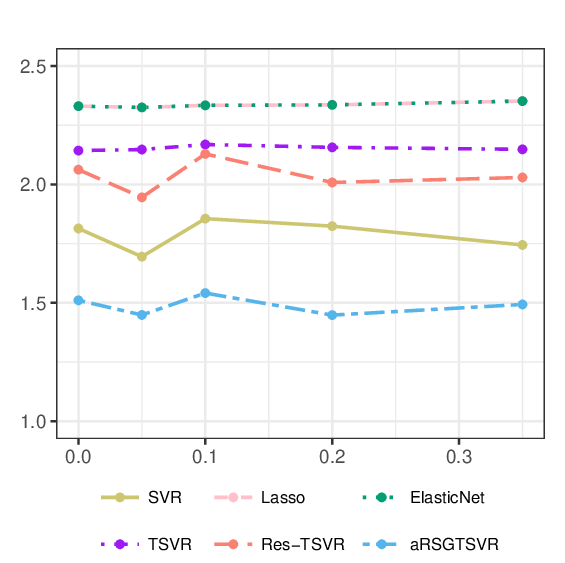}
	}
	\hfill
	\subfigure[$n=50,p=100$]{
		\includegraphics[width=0.31\textwidth]{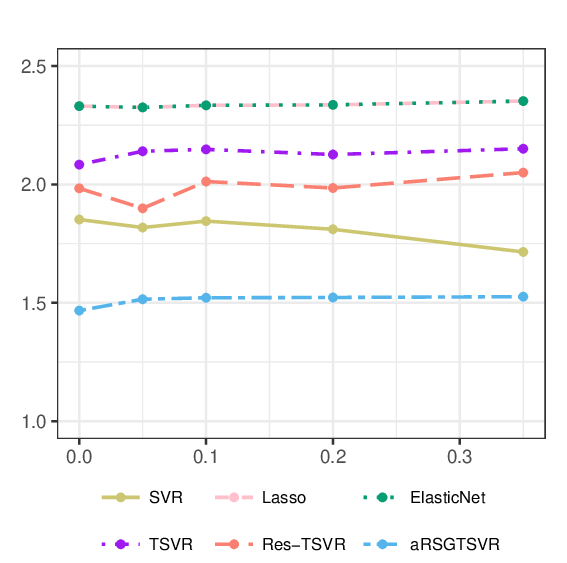}
	}
	\caption{Predictions of SVR, LASSO, Elastic Net, TSVR, Res-TSVR and aRSGTSVR on Sinc function with different types of noises.}
	\label{nonlinear 2.2}
\end{figure}

{ Table \ref{nonlinear 2.1} presents the RMSE and standard deviation results of all compared methods across different combinations of sample size $n$, feature dimension $p$, and label noise levels ranging from 0\% to 35\%, while Figure \ref{nonlinear 2.2} further visualizes the performance trends of each method under varying noise conditions for different dimensional settings. As shown in both the table and figure, the proposed aRSGTSVR consistently achieves the lowest RMSE values across all experimental scenarios, including low-dimensional, boundary high-dimensional, and classical high-dimensional cases, while maintaining stable performance with relatively small standard deviations. Compared with baseline models such as SVR, LASSO, Elastic Net, TSVR, and Res-TSVR, aRSGTSVR exhibits superior regression accuracy and stronger robustness against increasing label noise, confirming its combined advantages in prediction accuracy and stability for both low- and high-dimensional nonlinear regression tasks.}

\subsection{UCI datasets for classification}
\subsubsection{Experimental results}
To { assess the generalization performance} of the models, we conducted tests on a lot of UCI datasets (see Table \ref{uci-info}). Furthermore, to examine the robustness of the SVMs to outliers, we introduced artificial noise by randomly selecting 15\% and 35\% of the training samples and swapping their labels.

Table \ref{uci-00} presents the performance comparison of various SVM models using the initial data. In terms of average prediction accuracy, our proposed aRSGTSVM achieves the most optimal results, followed by RoBoSS-SVM, while rhingeSVM also demonstrates strong competitiveness. Although 1-SVM performs well only on certain datasets, all models can effectively predict data labels. TPMSVM and Pin-TSVM show the relatively poorest performance. Overall, the vast majority of models are capable of effectively classifying the data.

As shown in Table \ref{uci-15}, the aRSGTSVM model still achieves the optimal performance. In comparison, the competitive advantages of rhingeSVM, RoBoSS-SVM, and 1-SVM models gradually diminish, while TPMSVM shows an improving trend. Pin-TSVM consistently demonstrates relatively poor performance. From Table \ref{uci-35}, it can be observed that aRSGTSVM performs best on all UCI datasets, and as the noise ratio increases, the superiority of aRSGTSVM becomes more pronounced, indicating that our model exhibits strong robustness to label noise.

\begin{table}[!t]
	\centering
	\footnotesize
	\caption{The information of seleced UCI datasets.}
	\label{uci-info}
	\begin{tabularx}{0.6\textwidth}{XXX}
			\toprule
			Dataset & \#Samples & \#Features \\
			\midrule
			acoustic & 400 & 50 \\

			amphibians & 189 & 21 \\

			autism & 702 & 15 \\

			bads & 1372 & 4 \\

			dccc & 30000 & 23 \\

			diabetes & 520 & 16\\

			gait & 47 & 321 \\

			garments & 1197 & 7 \\

			heartfailure & 299 & 12 \\

			lsvt & 126 & 310 \\

			messidor & 1151 & 19 \\

			raisin & 900 & 7 \\

			shillbidding & 6321 & 9 \\

			tripadvisor & 980 & 9 \\

			vertebral & 310 & 6 \\

			waveform & 5000 & 40\\

			wp & 4746 & 15\\
			\bottomrule
	\end{tabularx}
\end{table}

\begin{table}[!t]
	\centering
	\caption{The results of UCI datasets with 0\% label noise. The bold is the best one.}
	\label{uci-00}
	\resizebox{\textwidth}{!}{
		\begin{tabular}{lllllll}
			\toprule
			\textbf{} & \textbf{1-SVM} & \textbf{TPMSVM} & \textbf{Pin-TSVM} & \textbf{rhingeSVM} & \textbf{\text{RoBoSS}-\text{SVM}} & \textbf{aRSGTSVM} \\
			\midrule
			acoustic & 0.755$\pm$0.084 & 0.825$\pm$0.062 & 0.767$\pm$0.059 & 0.868$\pm$0.044 & 0.853$\pm$0.046 & \pmb{0.873}$\pm$\pmb{0.022}\\

			amphibians & 0.704$\pm$0.080 & 0.746$\pm$0.090 & 0.714$\pm$0.115 & 0.793$\pm$0.062 & 0.809$\pm$0.075 & \pmb{0.814$\pm$0.083}\\

			autism & \pmb{1.000$\pm$0.000} & 0.967$\pm$0.030 & 0.970$\pm$0.038 & \pmb{1.000$\pm$0.000} & \pmb{1.000$\pm$0.000} & \pmb{1.000$\pm$0.000}\\

			bads & 0.981$\pm$0.017 & 0.985$\pm$0.012 & 0.961$\pm$0.026 & 0.987$\pm$0.009 & 0.984$\pm$0.013 & \pmb{0.988$\pm$0.013}\\

			dccc & 0.753$\pm$0.064 & 0.821$\pm$0.053 & 0.812$\pm$0.049 & 0.803$\pm$0.054 & 0.832$\pm$0.059 & \pmb{0.836$\pm$0.031}\\

			diabetes & 0.917$\pm$0.036 & 0.904$\pm$0.037 & 0.892$\pm$0.030 & 0.937$\pm$0.041 & \pmb{0.940$\pm$0.039} & 0.910$\pm$0.026\\

			gait & 0.980$\pm$0.063 & 0.915$\pm$0.145 & 0.915$\pm$0.145 & 0.980$\pm$0.063 & 0.940$\pm$0.135 & \pmb{1.000$\pm$0.000}\\

			garments & 0.999$\pm$0.003 & 0.998$\pm$0.004 & 0.993$\pm$0.008 & \pmb{1.000$\pm$0.000} & \pmb{1.000$\pm$0.000} & 0.999$\pm$0.003\\

			heartfailure & 0.762$\pm$0.071 & 0.836$\pm$0.058 & 0.843$\pm$0.052 & 0.833$\pm$0.079 & 0.836$\pm$0.060 & \pmb{0.859$\pm$0.063}\\

			lsvt & \pmb{0.985$\pm$0.032} & 0.760$\pm$0.123 & 0.824$\pm$0.094 & 0.953$\pm$0.075 & 0.904$\pm$0.123 & 0.912$\pm$0.090\\

			messidor & 0.600$\pm$0.051 & 0.699$\pm$0.042 & 0.601$\pm$0.071 & \pmb{0.751$\pm$0.036} & 0.673$\pm$0.060 & \pmb{0.751$\pm$0.034}\\

			raisin & 0.788$\pm$0.034 & 0.865$\pm$0.028 & 0.862$\pm$0.031 & 0.871$\pm$0.032 & \pmb{0.875$\pm$0.027} & 0.869$\pm$0.027\\

			shillbidding & 0.982$\pm$0.012 & 0.984$\pm$0.008 & 0.979$\pm$0.010 & 0.984$\pm$0.011 & 0.982$\pm$0.006 & \pmb{0.988$\pm$0.009}\\

			tripadvisor & 0.644$\pm$0.027 & 0.747$\pm$0.020 & 0.753$\pm$0.019 & 0.733$\pm$0.026 & 0.759$\pm$0.033 & \pmb{0.761$\pm$0.034}\\

			vertebral & 0.813$\pm$0.052 & 0.803$\pm$0.051 & 0.806$\pm$0.059 & 0.855$\pm$0.061 & \pmb{0.858$\pm$0.046} & 0.839$\pm$0.061\\

			waveform & 0.832$\pm$0.047 & 0.866$\pm$0.034 & 0.836$\pm$0.028 & 0.877$\pm$0.021 & 0.881$\pm$0.031 & \pmb{0.887$\pm$0.025}\\

			wp & 0.809$\pm$0.044 & 0.853$\pm$0.030 & 0.853$\pm$0.028 & 0.856$\pm$0.031 & 0.857$\pm$0.029 & \pmb{0.861$\pm$0.032}\\
			\bottomrule
		\end{tabular}
	}
\end{table}

\begin{table}[!t]
	\centering
	\caption{The results of UCI datasets with 15\% label noise. The bold is the best one.}
	\label{uci-15}
	\resizebox{\textwidth}{!}{
		\begin{tabular}{lllllll}
			\toprule
			\textbf{} & \textbf{1-SVM} & \textbf{TPMSVM} & \textbf{Pin-TSVM} & \textbf{rhingeSVM} & \textbf{\text{RoBoSS}-\text{SVM}} & \textbf{aRSGTSVM} \\
			\midrule
			acoustic & 0.670$\pm$0.079 & 0.760$\pm$0.079 & 0.755$\pm$0.077 & 0.767$\pm$0.060 & 0.780$\pm$0.088 & \pmb{0.850}$\pm$\pmb{0.057}\\

			amphibians & 0.688$\pm$0.075 & 0.730$\pm$0.067 & 0.730$\pm$0.114 & 0.778$\pm$0.077 & 0.740$\pm$0.078 & \pmb{0.784$\pm$0.106}\\

			autism & 0.829$\pm$0.127 & 0.959$\pm$0.025 & 0.953$\pm$0.025 & 0.949$\pm$0.024 & 0.963$\pm$0.026 & \pmb{0.992$\pm$0.015}\\

			bads & 0.817$\pm$0.032 & 0.970$\pm$0.027 & 0.955$\pm$0.025 & 0.973$\pm$0.024 & 0.977$\pm$0.016 & \pmb{0.978$\pm$0.020}\\

			dccc & 0.628$\pm$0.063 & \pmb{0.820$\pm$0.041} & 0.807$\pm$0.040 & 0.814$\pm$0.039 & 0.817$\pm$0.041 & 0.810$\pm$0.041\\

			diabetes & 0.691$\pm$0.084 & 0.908$\pm$0.031 & 0.902$\pm$0.044 & 0.869$\pm$0.039 & \pmb{0.917$\pm$0.016} & 0.904$\pm$0.035\\

			gait & 0.850$\pm$0.141 & 0.875$\pm$0.144 & 0.855$\pm$0.138 & 0.875$\pm$0.144 & 0.855$\pm$0.138 & \pmb{0.960$\pm$0.084}\\

			garments & 0.818$\pm$0.074 & 0.994$\pm$0.008 & 0.992$\pm$0.009 & 0.980$\pm$0.011 & \pmb{0.999$\pm$0.003} & 0.998$\pm$0.004\\

			heartfailure & 0.659$\pm$0.076 & 0.826$\pm$0.075 & 0.826$\pm$0.075 & 0.823$\pm$0.059 & 0.830$\pm$0.069 & \pmb{0.856$\pm$0.074}\\

			lsvt & 0.849$\pm$0.060 & 0.754$\pm$0.136 & 0.693$\pm$0.120 & 0.848$\pm$0.063 & 0.819$\pm$0.120 & \pmb{0.921$\pm$0.074}\\

			messidor & 0.561$\pm$0.042 & 0.698$\pm$0.045 & 0.642$\pm$0.035 & 0.699$\pm$0.040 & 0.646$\pm$0.053 & \pmb{0.731$\pm$0.051}\\

			raisin & 0.735$\pm$0.183 & 0.859$\pm$0.035 & 0.854$\pm$0.032 & 0.859$\pm$0.041 & 0.863$\pm$0.035 & \pmb{0.871$\pm$0.033}\\

			shillbidding & 0.903$\pm$0.058 & 0.973$\pm$0.012 & 0.968$\pm$0.022 & \pmb{0.978$\pm$0.011} & \pmb{0.978$\pm$0.010} & \pmb{0.978$\pm$0.013}\\

			tripadvisor & 0.601$\pm$0.035 & \pmb{0.752$\pm$0.035} & 0.747$\pm$0.035 & 0.731$\pm$0.041 & 0.750$\pm$0.042 & 0.739$\pm$0.044\\

			vertebral & 0.687$\pm$0.087 & 0.806$\pm$0.086 & 0.742$\pm$0.083 & 0.826$\pm$0.072 & \pmb{0.858$\pm$0.059} & \pmb{0.858$\pm$0.055}\\

			waveform & 0.707$\pm$0.050 & 0.815$\pm$0.028 & 0.828$\pm$0.032 & 0.843$\pm$0.030 & 0.831$\pm$0.034 & \pmb{0.860$\pm$0.032}\\

			wp & 0.669$\pm$0.043 & 0.827$\pm$0.031 & 0.818$\pm$0.026 & 0.824$\pm$0.030 & 0.827$\pm$0.026 & \pmb{0.831$\pm$0.033}\\
			\bottomrule
		\end{tabular}
	}
\end{table}

\begin{table}[!t]
	\centering
	\caption{The results of UCI datasets with 35\% label noise. The bold is the best one.}
	\label{uci-35}
	\resizebox{\textwidth}{!}{
		\begin{tabular}{lllllll}
			\toprule
			\textbf{} & \textbf{1-SVM} & \textbf{TPMSVM} & \textbf{Pin-TSVM} & \textbf{rhingeSVM} & \textbf{\text{RoBoSS}-\text{SVM}} & \textbf{aRSGTSVM} \\
			\midrule
			acoustic & 0.585$\pm$0.110 & 0.605$\pm$0.112 & 0.595$\pm$0.112 & 0.647$\pm$0.088 & 0.642$\pm$0.093 & \pmb{0.853$\pm$0.053}\\

			amphibians & 0.529$\pm$0.153 & 0.546$\pm$0.152 & 0.593$\pm$0.098 & 0.609$\pm$0.128 & 0.556$\pm$0.113 & \pmb{0.793$\pm$0.069}\\

			autism & 0.619$\pm$0.294 & 0.882$\pm$0.050 & 0.929$\pm$0.033 & 0.895$\pm$0.057 & 0.900$\pm$0.056 & \pmb{0.993$\pm$0.022}\\

			bads & 0.620$\pm$0.054 & 0.972$\pm$0.014 & 0.918$\pm$0.038 & 0.965$\pm$0.018 & 0.969$\pm$0.014 & \pmb{0.981$\pm$0.011}\\

			dccc & 0.570$\pm$0.113 & 0.778$\pm$0.056 & 0.549$\pm$0.110 & 0.787$\pm$0.047 & 0.779$\pm$0.043 & \pmb{0.799$\pm$0.050}\\

			diabetes & 0.610$\pm$0.171 & 0.810$\pm$0.066 & 0.835$\pm$0.052 & 0.821$\pm$0.048 & 0.835$\pm$0.047 & \pmb{0.902$\pm$0.032}\\

			gait & 0.755$\pm$0.227 & 0.640$\pm$0.251 & 0.685$\pm$0.221 & 0.680$\pm$0.258 & 0.620$\pm$0.253 & \pmb{0.980$\pm$0.063}\\

			garments & 0.705$\pm$0.348 & 0.973$\pm$0.015 & 0.985$\pm$0.013 & 0.969$\pm$0.023 & 0.989$\pm$0.011 & \pmb{0.999$\pm$0.003}\\

			heartfailure & 0.519$\pm$0.156 & 0.736$\pm$0.139 & 0.690$\pm$0.137 & 0.719$\pm$0.110 & 0.733$\pm$0.108 & \pmb{0.853$\pm$0.039}\\

			lsvt & 0.666$\pm$0.158 & 0.608$\pm$0.191 & 0.586$\pm$0.193 & 0.668$\pm$0.164 & 0.666$\pm$0.148 & \pmb{0.912$\pm$0.081}\\

			messidor & 0.554$\pm$0.045 & 0.617$\pm$0.052 & 0.557$\pm$0.063 & 0.590$\pm$0.071 & 0.602$\pm$0.065 & \pmb{0.717$\pm$0.035}\\

			raisin & 0.615$\pm$0.295 & 0.848$\pm$0.038 & 0.846$\pm$0.039 & 0.836$\pm$0.035 & 0.851$\pm$0.040 & \pmb{0.868$\pm$0.027}\\

			shillbidding & 0.632$\pm$0.073 & 0.973$\pm$0.018 & 0.975$\pm$0.016 & 0.970$\pm$0.016 & 0.971$\pm$0.017 & \pmb{0.981$\pm$0.014}\\

			tripadvisor & 0.535$\pm$0.046 & 0.668$\pm$0.084 & 0.602$\pm$0.179 & 0.728$\pm$0.058 & 0.714$\pm$0.071 & \pmb{0.742$\pm$0.053}\\

			vertebral & 0.635$\pm$0.171 & 0.716$\pm$0.087 & 0.603$\pm$0.143 & 0.707$\pm$0.062 & 0.687$\pm$0.063 & \pmb{0.855$\pm$0.060}\\

			waveform & 0.612$\pm$0.065 & 0.802$\pm$0.051 & 0.804$\pm$0.047 & 0.779$\pm$0.051 & 0.793$\pm$0.060 & \pmb{0.877$\pm$0.021}\\

			wp & 0.566$\pm$0.088 & 0.798$\pm$0.033 & 0.786$\pm$0.061 & 0.804$\pm$0.051 & 0.816$\pm$0.034 & \pmb{0.832$\pm$0.034}\\
			\bottomrule
		\end{tabular}
	}
\end{table}

\subsubsection{Comparisons by statistical test}
To further validate the effectiveness of our proposed aRSGTSVM model, we compare it with other SVM models on multiple datasets using the Friedman test and the corresponding Nemenyi post-hoc test\cite{demvsar2006statistical}. The null hypothesis of the Friedman test is that there is no significant difference among all models. If the null hypothesis is rejected, the Nemenyi test is conducted. The Friedman statistic is defined as:

\[
F_F = \frac{(D_n - 1)\chi^2_F}{D_n(C_k - 1) - \chi^2_F},
\]
where \( D_n \) is the number of datasets, and \( C_k \) is the number of models, 

\[
\chi^2_F = \frac{12D_n}{C_k(C_k + 1)} \left( \sum_{i=1}^{C_k} R_i^2 - \frac{C_k(C_k + 1)^2}{4} \right),
\]
where \( R_i \) is the average rank of the \(i\)-th model (calculated from the results in Table \ref{uci-00}-\ref{uci-35}).

Given $D_n = 17$ and $C_k = 6$, then for a noise ratio of 0\%, $F_F = 14.507$. For 15\%, $F_F = 27.788$. For 35\%, $F_F = 25.107$. In our experiment, \( F_F \) follows an \( F(5,80) \) distribution. At a significance level of 0.05, the critical value is $F_{0.05}(5,80) = 3.329$. Since \( F_F > 3.329 \), the null hypothesis of the Friedman test should be rejected, and the Nemenyi test is conducted next.

\begin{figure}[!t]
  \centering
  \subfigure[0\% label noise]{
  \begin{minipage}{0.95\linewidth}
      \centering
      \includegraphics[width=4in]{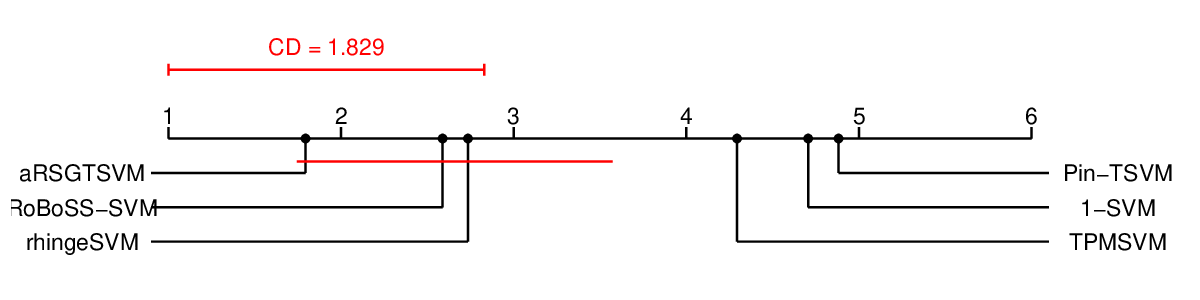}
  \end{minipage}}
  
  \subfigure[15\% label noise]{
  \begin{minipage}{0.95\linewidth}
      \centering
      \includegraphics[width=4in]{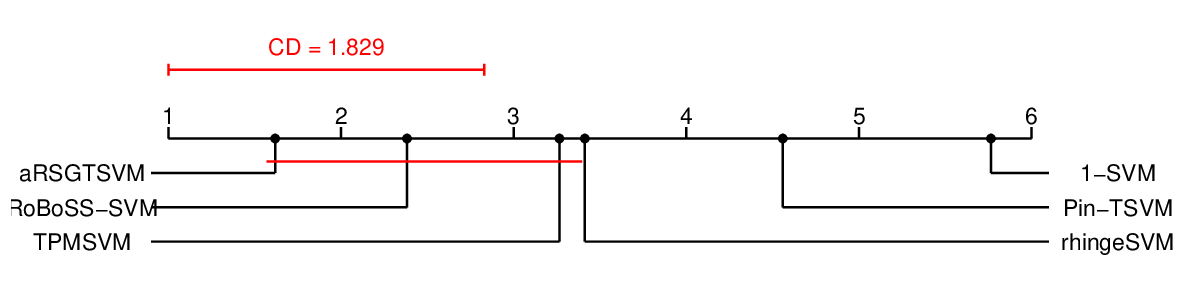}
  \end{minipage}}
  
  \subfigure[35\% label noise]{
  \begin{minipage}{0.95\linewidth}
      \centering
      \includegraphics[width=4in]{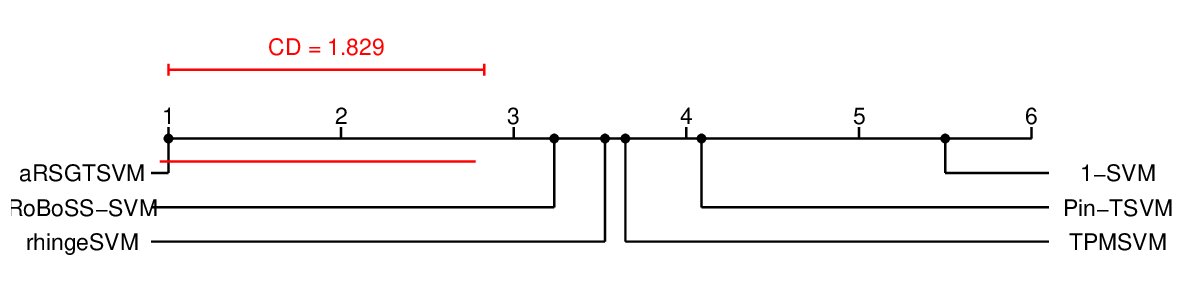}
  \end{minipage}}
  \caption{The average ranks of six classifiers on the UCI datasets with different label noise.}
  \label{fig:avg-rank}
\end{figure}

If there is a significant difference between two models, the difference in their average ranks must be at least the critical difference, defined as:  
\[CD = q_{0.05} \sqrt{\frac{C_k (C_k + 1)}{6D_n}}.\]  
Given that \( D_n = 17\), \( C_k = 6\), and \( q_{0.05}(6) =  2.85\), we obtain \( CD =  1.777\).

Figure \ref{fig:avg-rank} visualizes the average ranking comparison of six Support Vector Regression (SVR) models. According to the Critical Difference (CD) diagram analysis criterion, when the distance between two models on the horizontal axis exceeds the CD value (red solid line), it indicates a statistically significant difference between them. In the subfigure \ref{fig:avg-rank}(a), aRSGTSVM, rhingeSVM, and RoBoSS-SVM are grouped together, while Pin-TSVM, 1-SVM, and TPMSVM form another group, with significant performance differences between these two groups. In the subfigure \ref{fig:avg-rank}(b), the grouping changes to aRSGTSVM, RoBoSS-SVM, and TPMSVM as one group, and 1-SVM, Pin-TSVM, and rhingeSVM as the other group, maintaining significant inter-group differences. In the subfigure \ref{fig:avg-rank}(c), aRSGTSVM stands alone as one group, while the remaining models (RoBoSS-SVM, rhingeSVM, 1-SVM, Pin-TSVM, and TPMSVM) form another group, with equally significant performance differences between groups. It can be observed that aRSGTSVM significantly outperforms other comparative models, and its advantage becomes more pronounced as the noise ratio increases, fully demonstrating the model's performance stability across different label noise scenarios.

\subsection{Index tracking for regression}
In this section, { to assess the model’s out-of-sample performance}, we employ six index tracking datasets spanning from January 1, 2025, to July 1, 2025, for experimentation. The dataset is partitioned such that the first 70\% (rounded down if not an integer) is used as the training set, with the remainder serving as the test set. During training, the Mean Absolute Error (MAE) is utilized to select the optimal parameters, while the model's performance is evaluated based on the annual tracking error. The annual tracking error is defined as:

\[
\text{TrackingError}_{\text{year}} = \sqrt{252} \cdot \sqrt{\frac{\sum_{t=1}^{T}\left(err_t - \overline{err}\right)^2}{T-1}},
\]
where \(err_t = y_t - \hat{y}_t\) for \(t = 1, 2, \cdots, T\), \(y_t\) denotes the return of an index at time \(t\), \(\hat{y}_t\) is the estimated value of \(y_t\), and \(\overline{err}\) represents the mean of \(err_t\) over \(t = 1, 2, \cdots, T\).

\begin{table}[!t]
	\centering
	\scriptsize
	\caption{TrackingError$_{Year}$ for different datasets. The bold is the best one.}
	\label{index tracking1}
		\begin{tabularx}{\textwidth}{XXXXXXX}
			\toprule
			\textbf{Index} & \textbf{SVR} & \textbf{LASSO } & \textbf{Elastic Net} & \textbf{TSVR} & \textbf{Res-TSVR} & \textbf{aRSGTSVR} \\
			\midrule
			bz50 & 0.138 & 0.126 & 0.133 & 0.120 & 0.135 & \textbf{0.117}\\
			cy200 & 0.059 & 0.134 & 0.078 & 0.045 & 0.045 & \textbf{0.040}\\			
			hs300 & 0.277 & 0.277 & 0.380 & 0.122 & 0.138 & \textbf{0.117}\\
			xf100 & 0.638 & 0.328 & 0.250 & 0.368 & 0.925 & \textbf{0.226}\\
			ys50 & 0.191 & 0.065 & 0.068 & 0.128 & 0.124 & \textbf{0.057}\\
			zz500 & 0.297 & 0.271 & 0.185 & 0.162 & 0.109 & \textbf{0.105}\\
			\bottomrule
		\end{tabularx}
\end{table}

\begin{figure}[!t]
	\centering
	\subfigure[bz50\label{subfig:bz50}]{
		\includegraphics[width=0.31\textwidth]{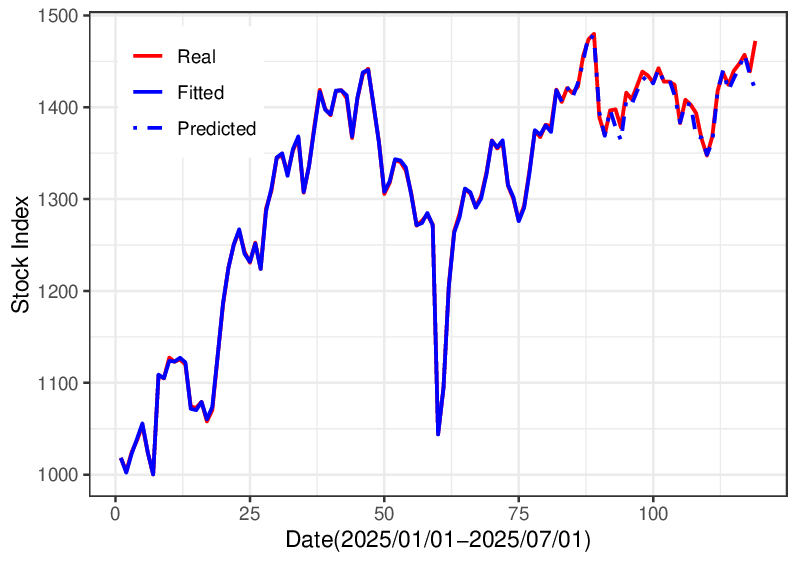}
	}\hfill
	\subfigure[cy200\label{subfig:cy200}]{
		\includegraphics[width=0.31\textwidth]{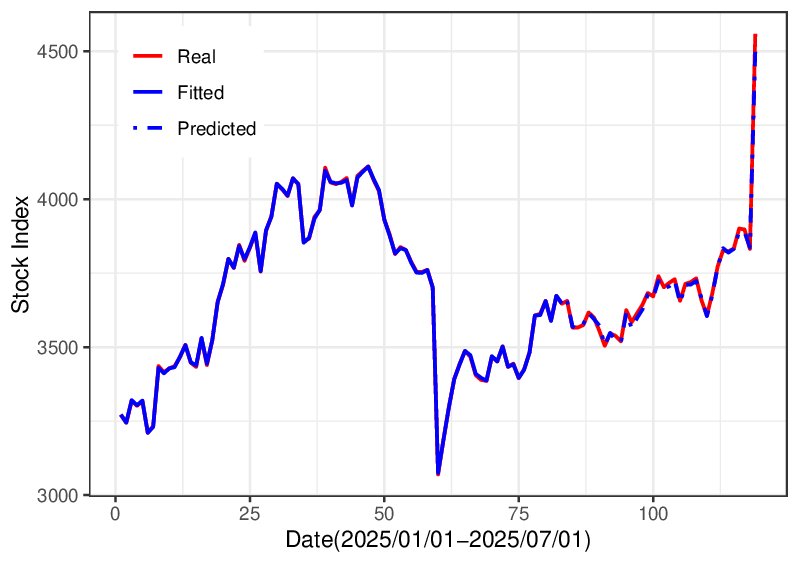}
	}\hfill
	\subfigure[hs300\label{subfig:hs300}]{
		\includegraphics[width=0.31\textwidth]{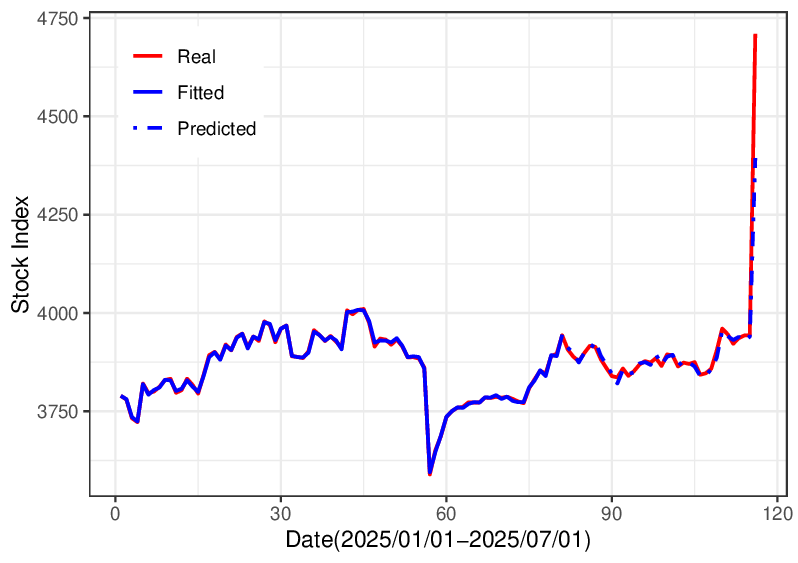}
	}\hfill
	\subfigure[xf100\label{subfig:xf100}]{
		\includegraphics[width=0.31\textwidth]{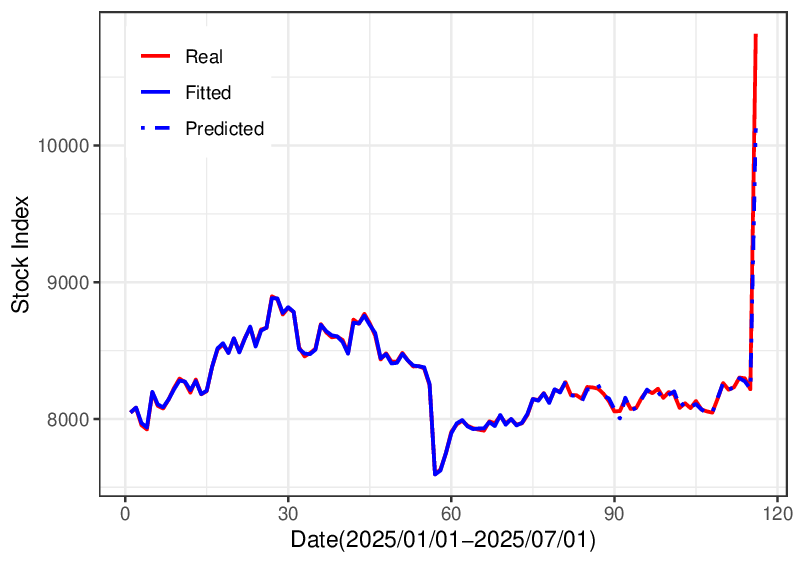}
	}\hfill
	\subfigure[ys50\label{ys50}]{
		\includegraphics[width=0.31\textwidth]{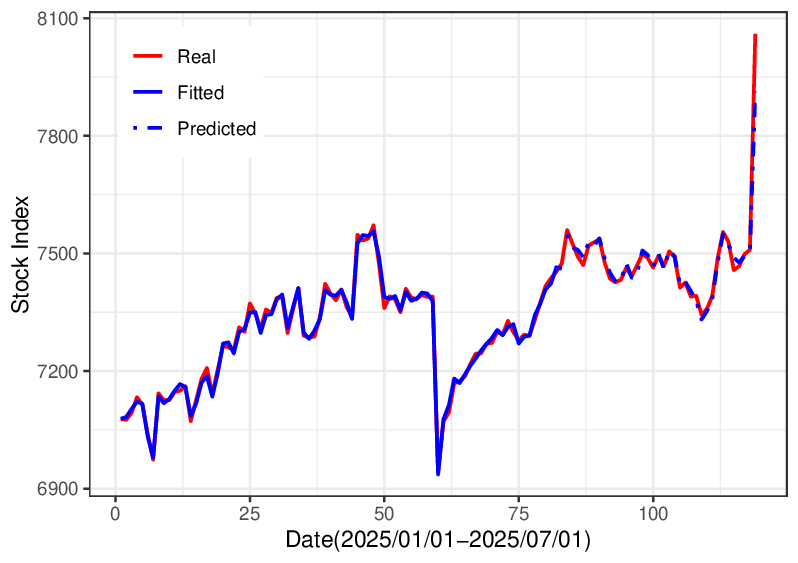}
	}\hfill
	\subfigure[zz500\label{subfig:zz500}]{
		\includegraphics[width=0.31\textwidth]{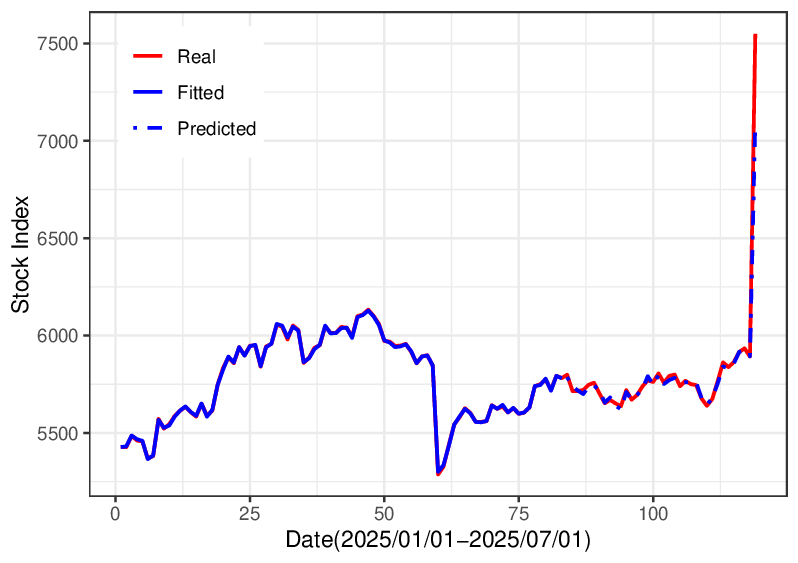}
	}
	\caption{
		Index tracking plots for Stock Indices. The red solid lines are the true values, the blue solid lines are the fitted values on the training set, and the blue dashed lines are the predicted values on the test set.}
	\label{index tracking2}
\end{figure}

To mitigate spurious regression and enable the model to focus on replicating relative price movements, converting raw price data to returns is crucial for index tracking. For a raw price series \( P_t \), the return at time \( t \) is calculated as:

\[
r_t = \frac{P_t}{P_{t-1}} - 1, \quad t = 1, 2, \cdots, T.
\]

Similarly, the index value \( y_t \) at time \( t \) is converted to its corresponding return, which we continue to denote as \( y_t \) for consistency.

The experimental results are presented in Table \ref{index tracking1}. As evidenced by the annual tracking errors, the proposed aRSGTSVR achieves the best performance across all six index tracking datasets, demonstrating strong generalization capability of the model. As illustrated in Figure \ref{index tracking2}, the fitted and predicted values of the aRSGTSVR model closely align with the true values. This visually confirms the model's superior fitting performance and robust generalization ability.

\section{Conclusion}\label{sec:con}
{ This paper designs a novel asymmetric loss function \(L_{aR}\) based on the RoBoSS loss, which is integrated with ENNHSVM to establish aRSGTSVM for classification and aRSGTSVR for regression tasks. To tackle the challenging optimization issue arising from the combination of a nonconvex loss \(L_{aR}\) and a sparse penalty \(l_1\) in the objective function, the iPiano algorithm is adopted to achieve efficient and stable optimization. Experimental results in synthetic datasets demonstrate that aRSGTSVM maintains strong robustness against label noise, delivers stable performance under resampling, achieves satisfactory results in high-dimensional scenarios, and presents competitive computational efficiency. Meanwhile, aRSGTSVR also exhibits prominent noise resistance and effective high-dimensional variable selection capability. In real-world datasets, both proposed models acquire favorable generalization performance. The core contribution of this study lies in that the designed asymmetric loss naturally balances model robustness and sparsity, which provides a reliable and superior alternative to conventional symmetric loss functions for practical scenarios suffering from label noise, high-dimensional features, and resampling instability.}

 { It is noteworthy that although this study employs a cross-validation strategy for model parameter optimization, this method can only detect the presence or absence of overfitting rather than alleviate it. Alleviating overfitting requires strategies such as regularization. Furthermore, cross-validation necessitates repeated data partitioning and model retraining, resulting in substantial computational overhead. This issue becomes particularly pronounced in high-dimensional data or complex model scenarios. Given that information criteria (e.g., AIC, BIC) can construct objective functions by integrating model goodness of fit and complexity without requiring repeated data partitioning and retraining, they significantly enhance parameter tuning efficiency while balancing model generalization performance. Therefore, exploring the optimal information criterion tailored to the characteristics of our model represents a highly valuable direction for future research. Additionally, distributed or parallel algorithms can be employed to further improve the computational efficiency of the model.}

\section*{Acknowledgments}
This research is supported by the National Natural Science Foundation of China (Grant No. 12401664), the Natural Science Foundation of Chongqing Municipality (Grant No. CSTB2024NSCQ-MSX0855), and the Science and Technology Research Program of Chongqing Municipal Education Commission (Grant No. KJQN202400514).

\end{document}